\documentclass[preprint,12pt]{elsarticle}
\usepackage{booktabs}
\usepackage{multirow}
\usepackage{hyperref}
\usepackage[utf8]{inputenc}
\usepackage{amssymb}
\usepackage{amsmath}
\usepackage[T1]{fontenc}
\begin{document}

\begin{frontmatter}



\title{EndoChat: Grounded Multimodal Large Language Model for Endoscopic Surgery}

\author[label1,label2,†]{Guankun Wang}
\author[label1,label3,†]{Long Bai}
\author[label1,†]{Junyi Wang}
\author[label3,label4,†]{Kun Yuan}
\author[label5]{Zhen Li}
\author[label1]{Tianxu Jiang}
\author[label1]{Xiting He}
\author[label6]{Jinlin Wu}
\author[label6]{Zhen Chen}
\author[label6]{Zhen Lei}
\author[label6]{Hongbin Liu}
\author[label2]{Jiazheng Wang}
\author[label2]{Fan Zhang}
\author[label4]{Nicolas Padoy}
\author[label3]{Nassir Navab}
\author[label1,*]{Hongliang Ren}

\affiliation[label1]{organization={The Chinese University of Hong Kong},%Department and Organization
            city={Hong Kong SAR},
            postcode={999077}, 
            country={China}}
\affiliation[label2]{organization={Theory Lab, Central Research Institute, 2012 Labs, Huawei Technologies Co. Ltd.},%Department and Organization
            city={Hong Kong SAR},
            postcode={999077}, 
            country={China}}
\affiliation[label3]{organization={Chair of Computer Aided Procedures (CAMP), Technical University of Munich},%Department and Organization
            city={Munich},
            postcode={81927}, 
            country={Germany}}
\affiliation[label4]{organization={University of Strasbourg, CNRS, INSERM, ICube \& IHU Strasbourg},%Department and Organization
            city={Strasbourg},
            postcode={67200}, 
            country={France}}
\affiliation[label5]{organization={Department of Gastroenterology, Qilu Hospital of Shandong University},%Department and Organization
            city={Jinan},
            postcode={250000}, 
            country={China}}
\affiliation[label6]{organization={Centre for Artificial Intelligence and Robotics (CAIR), Hong Kong Institute of Science \& Innovation, Chinese Academy of Sciences},%Department and Organization
            city={Hong Kong SAR},
            postcode={999077}, 
            country={China}}

%% Abstract
\begin{abstract}
Recently, Multimodal Large Language Models (MLLMs) have demonstrated their immense potential in computer-aided diagnosis and decision-making. In the context of robotic-assisted surgery, MLLMs can serve as effective tools for surgical training and guidance. However, there is still a deficiency of MLLMs specialized for surgical scene understanding in endoscopic procedures. To this end, we present EndoChat MLLM to address various dialogue paradigms and subtasks in understanding endoscopic procedures. To train our EndoChat, we construct the Surg-396K dataset through a novel pipeline that systematically extracts surgical information and generates structured annotations based on large-scale endoscopic surgery datasets. Furthermore, we introduce a multi-scale visual token interaction mechanism and a visual contrast-based reasoning mechanism to enhance the model's representation learning and reasoning capabilities. Our model achieves state-of-the-art performance across five dialogue paradigms and seven surgical scene understanding tasks. Additionally, we conduct evaluations with professional surgeons, who provide positive feedback on the majority of conversation cases generated by EndoChat. Overall, these results demonstrate that EndoChat has the potential to advance training and automation in robotic-assisted surgery. Our dataset and model are publicly available at \href{https://github.com/gkw0010/EndoChat} {https://github.com/gkw0010/EndoChat}.
\end{abstract}


% \begin{highlights}
% \item Propose EndoChat for visual grounding conversations in endoscopic surgery.
% \item Develop Surg-396K, a multimodal surgical dataset with 396K image-instruction pairs.
% \item Integrate Mixed Visual Token Engine for enhanced visual feature extraction.
% %\item Introduce a visual contrast-based method to address object hallucinations.
% \item Achieve superior performance in surgical scene understanding dialogues and tasks.
% \item Positive feedback from surgeons indicates potential for training support.
% \end{highlights}

%% Keywords
\csname
\endcsname
\end{frontmatter}



\section{Introduction}
\label{sec:intro}
Robot-assisted surgery (RAS) offers unprecedented opportunities to enhance surgical precision, minimize patient trauma, and shorten postoperative recovery times~\cite{nwoye2023cholectriplet2021,nwoye2025surgitrack,islam2021st}. However, the effective application of this technology places significant demands on the skills of surgeons, particularly in mastering the operation of robotic systems during procedures~\cite{alabi2025multitask,wagner2023comparative}. To ensure surgical safety and efficacy, surgeons must undergo rigorous training to acquire the core skills required for robotic operation~\cite{chen2020comprehensive,aziz2021effect}. To improve the efficiency of this training process, various simulator-based surgical platforms~\cite{mariani2020skill} have been developed. However, when trainees encounter challenges during training, they often require immediate feedback and guidance from professional surgeons to resolve doubts or correct mistakes. Unfortunately, professional surgeons typically face significant time constraints due to their heavy workload in clinical, teaching, and research responsibilities, making it difficult for them to provide consistent, real-time support during training~\cite{sharma2021medfusenet,seenivasan2022surgical}. As a result, there is a pressing need for technological solutions that can deliver flexible, real-time, and efficient support in surgical training.

Recently, artificial intelligence (AI)-based dialogue systems with structured Visual Question Answering (VQA) have been introduced into surgical training~\cite{seenivasan2022surgical,bai2023surgical,he2024pitvqa}. These systems analyze visual data from surgical scenarios to address trainees' questions. However, their reliance on simple structured and object presence-based VQA datasets, typically trained for specific tasks, limits their ability to dynamically adapt to the wide range of questions posed by trainees~\cite{pellegrini2023rad,lin2023medical}. Moreover, most of these approaches are based on encoder-decoder architectures, which require explicitly defined input/output formats. This rigidity makes them less flexible for handling highly open-ended generative tasks and limits their scalability. When applied beyond their designed scope, these models often show significant performance drops, making them poorly suited for the complexity of diverse surgical scenarios~\cite{christopher2024machine,gozalo2023chatgpt}. When trainees ask open-ended questions, existing VQA systems lack the flexibility and contextual understanding needed for such interactions. As a result, they struggle to handle open-ended questions or complex multi-turn dialogues, significantly limiting their usefulness in surgical training.

\begin{figure*}
    \centering
    \includegraphics[width=0.98\textwidth, trim=0 0 0 0]{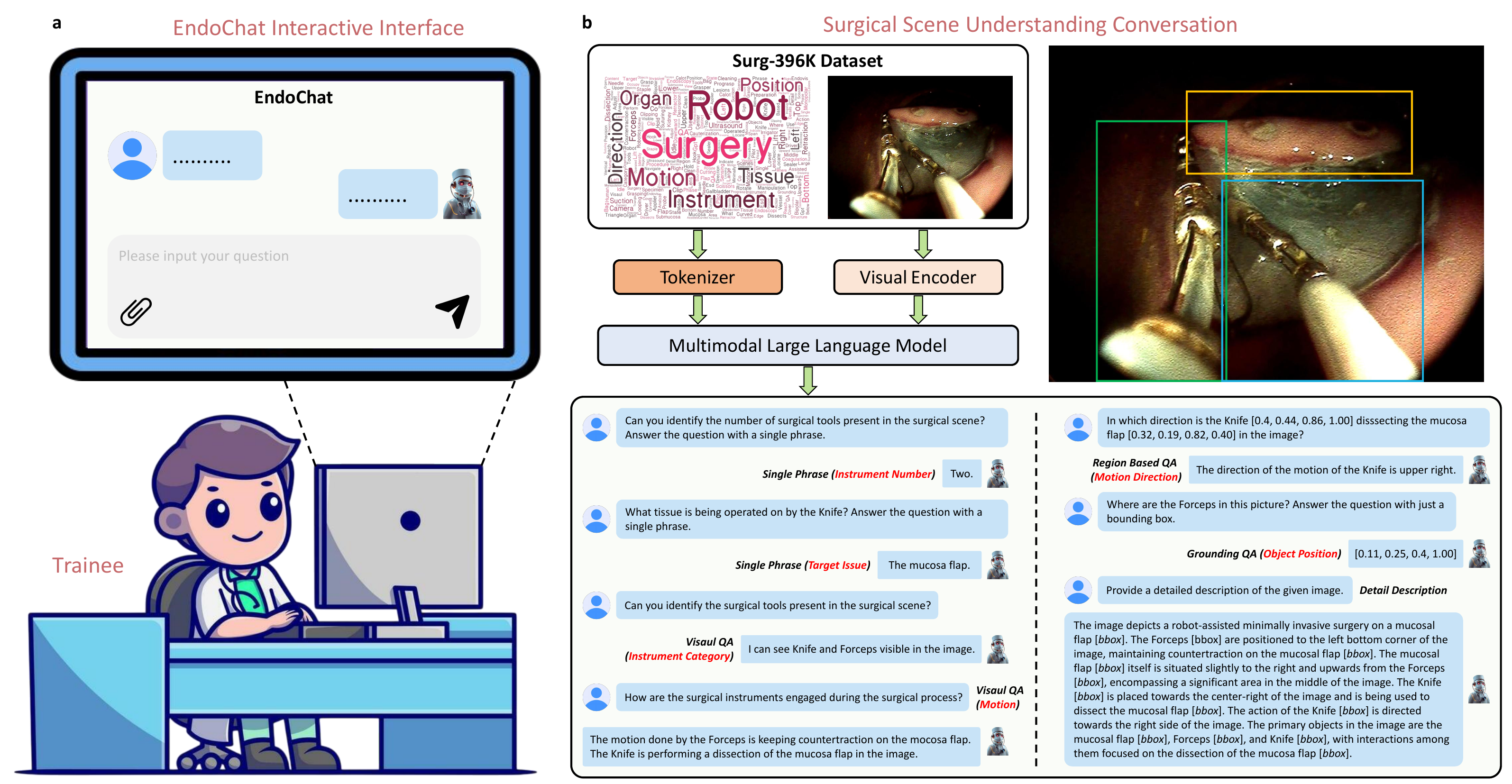}
    \caption{\textbf{Overview of the EndoChat}. \textbf{a} EndoChat is an interactive multimodal chatbot designed for surgical education and training. Users can interact with EndoChat by uploading images and formulating questions, enabling a comprehensive surgical scene understanding. \textbf{b} EndoChat is trained on Surg-396K, a large-scale multimodal instruction dataset. Surg-396K includes five conversation paradigms, enabling EndoChat to effectively perform natural language and visual grounding conversations with trainees. On the bottom is an example of the multi-turn conversation.}
    \label{fig:intro}
\end{figure*}

Currently, medical Multimodal Large Language Models (MLLMs) are emerging as a promising solution, offering significant potential through large-scale pretraining to perform complex reasoning and understanding across tasks~\cite{li2024llavamed,chen2024huatuogpt, chen2024gmai,liu2023medical}. Specifically, MLLMs can extract information from multimodal data in surgical scenarios and perform advanced reasoning. Unlike structured question-answering systems, MLLMs are capable of processing unstructured and complex contextual information. For example, trainees can ask open-ended questions in natural language, and MLLMs can generate targeted responses by leveraging their pre-trained knowledge and multimodal reasoning capabilities. These powerful natural language processing abilities enable MLLMs to handle multi-turn conversations and dynamically adjust responses based on context. This interaction is akin to receiving guidance from professional surgeons, significantly enhancing the training experience. Overall, MLLMs have the potential to partially replace guidance from professional surgeons by simulating their knowledge and decision-making abilities. This can address the limitations of current training solutions and alleviate the burden on surgeons who are constrained by demanding clinical schedules, ultimately improving the efficiency and quality of surgical training~\cite{chen2024huatuogpt,dou2023shennonggpt,he2024meddr}.

% Some prior works have explored the capabilities of MLLMs in surgical scenarios~\cite{li2024llava,jin2024surgicalllava,wang2024surgicallvlm}. However, we aim to further develop MLLMs to handle diverse queries from trainees in real-world scenarios. For example, a trainee might ask a brief question about a specific target, which does not require any redundant information. Therefore, \textit{Single Phrase QA} is designed to provide concise, accurate, and direct responses relevant to surgical contexts. When a trainee requests a detailed explanation of an entire surgical image, \textit{Detailed Description}, on the other hand, offers surgeons comprehensive explanations of all sub-tasks present within the current surgical scene. For routine queries about image content, \textit{Visual QA} delivers context-aware answers by combining user queries with image information. When a trainee needs a focused conversation about a specific region, \textit{Region Based QA} focuses on targeted responses for specific areas of interest, while \textit{Grounding QA} generates bounding box descriptions based on user-provided prompts. These five distinct conversational paradigms encompass the majority of scenarios encountered in natural language dialogues~\cite{kuckreja2024geochat}. By constructing them, we aim to design a surgical MLLM that can respond appropriately in different cases, creating a system that is better suited for human interaction. This would provide more practical and effective support for surgical training and education.

Surgical MLLMs have attracted significant attention from researchers~\cite{li2024llava,jin2024surgicalllava,wang2024surgicallvlm}. However, existing surgical MLLMs primarily focus on one of three approaches: leveraging the powerful capabilities of MLLMs to enhance question-answering performance~\cite{hou2024memory}, extending models for specific predefined tasks through instruction tuning~\cite{wang2024surgicallvlm}, or using web-sourced data to construct general-purpose descriptive and conversational datasets~\cite{li2024llava}. These methods face two major challenges: (i) In real-world scenarios, the queries from trainees are highly diverse. Solely relying on predefined query formats or generic captioning is insufficient to handle the variety of queries from trainees in practical settings. (ii) Many current MLLM applications rely on pre-trained visual encoders to extract visual features~\cite{li2024llava,liu2024visual}. However, due to the domain gap, semantic information from general-purpose visual scenes cannot be directly applied to surgical scenarios. This leads to an inadequate understanding of visual information and the occurrence of reasoning hallucinations.  

To this end, we address the above challenges from two perspectives: constructing open-ended, knowledge-required surgical vision-language datasets and developing vision-enhanced MLLM models tailored for surgical scenarios. 
% Firstly, we develop Surg-396K, a surgical multimodal instruction dataset specifically for surgical contexts. We systematically extract surgical information from four public datasets by designing surgical attributes, and generate instruction-tuning data using diverse conversational templates. To better simulate real-world scenarios, we design five distinct conversational paradigms that encompass the majority of scenarios encountered in natural language dialogues~\cite{kuckreja2024geochat}, including \textit{Single Phrase QA}, \textit{Detailed Description}, \textit{Visual QA}, \textit{Region Based QA}, and \textit{Grounding QA}. Surg-396K provides comprehensive coverage of downstream tasks and conversational paradigms across a wide range of surgical scenarios. Therefore, the MLLM trained on Surg-396K is better suited for surgeon-system interaction and can respond appropriately in surgical situations, providing more practical and effective support for surgical training and education.
Firstly, we develop Surg-396K, a surgical multimodal instruction dataset specifically for surgical contexts. We systematically extract surgical attribute information from three public datasets and generate instruction-tuning data using diverse conversational templates. To better simulate real-world scenarios, as presented in Table~\ref{tab:data}, we propose five distinct conversational paradigms to capture the majority of natural language dialogue scenarios and define seven attribute-specific, surgery-related downstream tasks to ensure comprehensive coverage of surgical scene understanding. As a result, the MLLM trained on Surg-396K is better equipped for surgeon-system interaction, enabling it to respond effectively in surgical contexts and provide practical, reliable support for surgical training and education.
Secondly, in the model architecture of MLLMs, we design the Mixed Visual Token Engine (MVTE) to extract visual information at multiple scales for better vision-language alignment in surgical contexts. Unlike traditional frameworks that directly use pre-trained Vision Transformers (ViTs) to extract visual tokens, MVTE uses multiple visual towers to extract, interact, and fuse visual tokens, improving the visual information extraction before aligning with text. 
% Furthermore, to reduce model hallucinations and enhance scene understanding, we further introduce a visual contrast-based method to ensure consistency between visual inputs and language outputs in complex endoscopic surgery scenarios.
Furthermore, to minimize model hallucinations, we propose a visual contrast-based approach that compares outputs generated from both original and distorted visual inputs. This method refines the token selection process by applying adaptive plausibility constraints, ensuring better consistency between visual inputs and language outputs in complex endoscopic surgery scenarios
% Lastly, we conduct extensive experimental evaluations across five conversational paradigms and seven surgical scene understanding tasks. We also collaborate with professional surgeons to assess the usability of our proposed approach. The outstanding experimental results demonstrate that our dataset and model can serve as a powerful assistant for the comprehensive understanding of surgical scenarios.
%Moreover, we introduce a visual contrast mechanism to mitigate object hallucinations, further enhancing the model's visual understanding and ensuring consistency between visual inputs and language outputs in complex endoscopic surgery scenarios.

As a result, we propose EndoChat (shown in Figure~\ref{fig:intro}) to support diverse conversational paradigms in endoscopic surgical scenarios. This flexible framework effectively addresses diverse interaction needs and supports a wide range of surgical tasks, making it highly adaptable to the varied questions that trainees may pose in different contexts.
% As a result, we propose EndoChat (shown in Figure~\ref{fig:intro}) to support diverse conversational paradigms in endoscopic surgical scenarios. This flexible framework addresses various interaction needs and accommodates a wide range of surgical tasks such as instrument recognition, motion recognition, target localization, tissue identification, instrument counting, and motion direction detection, which is summarized based on our designed surgical attributes and highly adaptable to various questions that the trainee might raise in different contexts. 
To validate the effectiveness of EndoChat, we first conduct rigorous comparisons with commercial and open-source MLLMs across different dialogue paradigms. The results demonstrate that our approach surpasses existing general-purpose and medical MLLMs in terms of both surgical understanding accuracy and dialogue capability. Furthermore, we show that our model achieves state-of-the-art performance across various attribute-related sub-tasks of surgical scene understanding. Ablation studies further confirm the effectiveness of our innovative architectural design within the MLLM framework.
In addition, we invite experienced practicing surgeons to independently evaluate whether the assistant is beneficial for advancing surgical training procedures and whether they would be willing to adopt it. The evaluation results indicate that surgeons hold a positive attitude toward our proposed EndoChat, further demonstrating that EndoChat is a qualified assistant for various surgical training and education scenarios.
In summary, EndoChat marks a notable advancement in applying MLLMs to surgical training, delivering intelligent, context-aware assistance to trainees. The key contributions of this work are summarized as follows: 
\begin{itemize} 
\item[--] We propose EndoChat, a novel grounded MLLM that supports five conversational paradigms and seven attribute-related surgical sub-tasks in endoscopic surgical scenarios, addressing the requirement for effective dialogue systems in surgical training and guidance.
\item[--] We develop Surg-396K, a comprehensive multimodal surgical dataset containing 396K image-instruction pairs through a multi-conversation construction pipeline that systematically extracts surgical information and generates structured annotations.
\item[--] EndoChat incorporates the Mixed Visual Token Engine that enhances multi-scale visual information extraction and fusion. Additionally, a visual contrast-based method is integrated to address object hallucinations within MLLMs.
\item[--] Extensive experiments on our proposed dataset demonstrate that EndoChat outperforms existing general-purpose and medical MLLMs across various dialogue paradigms and surgical scene understanding sub-tasks. We also validate the practical efficacy of EndoChat through expert evaluations by experienced endoscopists, confirming its potential as an effective tool for enhancing surgical training and education. 
\end{itemize}

%To validate the effectiveness of EndoChat, we first conducted comparisons with open-source MLLMs across different dialogue paradigms. The results demonstrate that our approach surpasses existing general-purpose and medical MLLMs in terms of both surgical understanding accuracy and dialogue capability. Furthermore, we show that our model achieves state-of-the-art performance across various attribute-related sub-tasks of surgical scene understanding. Ablation studies further confirm the effectiveness of our innovative architectural design within the MLLM framework.
%Moreover, we invite experienced practicing surgeons to independently evaluate EndoChat's utility for advancing surgical training and their willingness for using it. The positive feedback indicate that EndoChat is a qualified assistant for various surgical training and education scenarios.

\section{Related work}
\subsection{Multimodal Large Language Models}
Recent studies, such as GPT-4V~\cite{achiam2023gpt} and Qwen-VL~\cite{wang2024qwen2}, have shown that large language models (LLMs) can effectively process visual information and generate corresponding textual descriptions. These advancements have sparked significant interest in MLLM research. The architectures of current MLLMs are relatively standardized, typically comprising four main components: a visual encoder, a text tokenizer, an alignment module, and an LLM~\cite{wang2024qwen2,li2023blip,xue2024xgen,zhu2023minigpt4,chen2023minigptv2}. The visual encoder uses pre-trained vision models to transform images into tokens that are interpretable by LLMs. Commonly employed vision models include CLIP-ViT~\cite{radford2021learning}, DINOv2~\cite{oquab2023dinov2}, and InternViT~\cite{chen2024far}. Similarly, the text tokenizer utilizes methods like Byte Pair Encoding (BPE)~\cite{shibata1999byte} or WordPiece~\cite{song2020fast} to convert textual inputs into tokenized representations.
Before integrating visual and textual tokens into the LLM, a vision-language alignment process is necessary to enable effective multimodal semantic understanding. Several studies have explored alignment strategies, including the Perceiver Resampler from Flamingo~\cite{alayrac2022flamingo}, the Q-Former from BLIP-2~\cite{li2023blip}, and the simple linear layer of LLaVA~\cite{liu2024improved,liu2024visual}, to enhance the model's capabilities of attending to visual information conditioned on text prompts. After alignment, the vision-language embeddings are passed into the LLM, which uses self-attention mechanisms and autoregressive language modeling to generate the final textual outputs~\cite{vaswani2017attention,brown2020language,yang2025qwen2}.
Furthermore, training MLLMs requires large-scale vision-language paired datasets~\cite{zhou2025tau,kuang2024natural}. To address the challenge of data scarcity, the LLaVA models employed visual instruction tuning and leveraged the advanced visual understanding capabilities of commercial MLLMs to extract diverse vision-language paired datasets~\cite{liu2024visual,liu2024improved}. This approach significantly enriched the open-source community by contributing foundational data resources.

Based on the strong visual understanding capabilities of MLLMs in general scenarios, researchers have developed powerful multimodal medical assistants~\cite{lu2025integrating,wang2023xrayglm,zhou2025training,liu2023medical,song2024pneumollm}. Early attempts focused on dataset-level efforts, such as collecting and organizing new medical datasets and performing text augmentation through instruction tuning~\cite{tu2024towards,li2024llavamed}. In addition, common computer-assisted diagnosis tasks have been widely explored, including image diagnosis for X-rays~\cite{wang2024interactive} and dermatological diagnosis tasks~\cite{zhou2024pre}. 
For medical report generation tasks, XrayGPT utilized advanced vision-language models to produce interactive summaries from radiology reports, offering concise findings and supporting follow-up questions~\cite{thawakar2024xraygpt}. Huang et al. aimed to refine medical reports by focusing on key semantic information, which enhanced the accuracy and interpretability of the generated content~\cite{huang2024enhancingmia}. Other studies have worked on improving classification tasks by integrating privacy-preserving LLMs and multi-type annotations into datasets, helping address the challenge of noisy labels~\cite{lanfredi2025enhancing}. However, due to the limited availability of high-quality annotations in endoscopic surgery and the significant domain gap between endoscopic surgery and general scenarios, the MLLM applications in the surgical field, while having seen some initial exploration, still face substantial challenges.

\subsection{Surgical Vision-Language Models}

Early surgical vision-language models primarily focused on developing VQA systems to address question-answering and dialogue requirements during surgical procedures~\cite{bai2023catvil,bai2023revisiting,chen2024llm,he2024pitvqa,peng2024prior,zhu2024alignment}. As the first approach to introduce a question-answering model specifically tailored for surgical scenarios, Surgical-VQA represented key elements in surgical environments—such as instruments, tissues, tools, and spatial positions—using textual descriptions~\cite{seenivasan2022surgical}. Built on the VisualBERT~\cite{li2019visualbert} framework, it integrated multimodal representations of text and images to generate corresponding answers through a decoder. Subsequent works, such as Surgical-VQLA, improved upon this method by incorporating bounding box outputs at the decoder stage, enabling explicit visual localization to better assist surgeons~\cite{bai2023catvil,bai2023surgical,he2024pitvqa,zhu2024alignment}. SSG-VQA addressed the shortage of vision-language datasets in surgical scenarios by generating synthetic surgical dialogue data through predefined attributes and templates.
Later advancements explored alternative network architectures~\cite{zhang2024dual,bai2025surgicalvqla,seenivasan2023surgicalgpt} and optimal vision-language alignment paradigms~\cite{yuan2024hecvl, yuan2023learning, yuan2025procedure}. However, as discussed in Section~\ref{sec:intro}, structured dialogue datasets combined with encoder-decoder architectures limit the model's capabilities, confining it to queries within predefined content. This restricts its ability to handle complex interactions, thereby reducing its clinical applicability.

With the rise of MLLMs~\cite{wang2024qwen2,achiam2023gpt,liu2024improved,liu2024visual}, the medical community is beginning to explore their potential applications in the field of healthcare~\cite{li2024llavamed,chen2024huatuogpt,li2024gmai,chen2024gmai, hu2024omnimedvqa, ferber2024context}. Several prior studies have demonstrated the potential of MLLMs in surgical scenarios~\cite{li2024llava,jin2024surgicalllava,wang2024surgicallvlm,schmidgall2024gpvls,hou2024memory}. For example, both Surgical-LVLM~\cite{wang2024surgicallvlm} and Surgical-LLaVA~\cite{jin2024surgicalllava} extended the instruction-tuning framework of LLaVA~\cite{liu2024visual} by incorporating existing classification annotations or structured texts into the training of MLLMs. Moreover, Surgical-LVLM integrated an additional localization module, enabling the generation of bounding boxes for specific targets. SCAN~\cite{hou2024memory} introduced a memory-augmented query mechanism to enhance the VQA performance by employing self-contained queries within the MLLM. LLaVA-Surg collected and annotated open-source videos and datasets to improve the model’s conversational capabilities through instruction tuning~\cite{li2024llava}. In contrast, GP-VLS unified various surgical tasks within a question-answering framework, representing the outputs of different downstream surgical tasks in textual form~\cite{schmidgall2024gpvls}. CoPESD introduced a novel surgical scene understanding dataset. After annotating the instrument-tissue-action information, it further combined instruction tuning and professional surgeons' descriptions to build a multi-granularity surgical motion analysis dataset~\cite{wang2024copesd}. However, existing methods are mostly based on general captioning or predefined surgical scene understanding tasks (e.g., action analysis, instrument recognition) and aim to enhance dialogue diversity through techniques such as instruction tuning. These approaches overlook the dialogue paradigm required for real-world interactions with surgeons, which is the key issue we aim to address in this paper.

\section{Methods}

\subsection{Surgical Multimodal Instruction Dataset: Surg-396K}
%数据集概述

\begin{figure*}
    \centering
    \includegraphics[width=0.98\textwidth, trim=0 0 0 0]{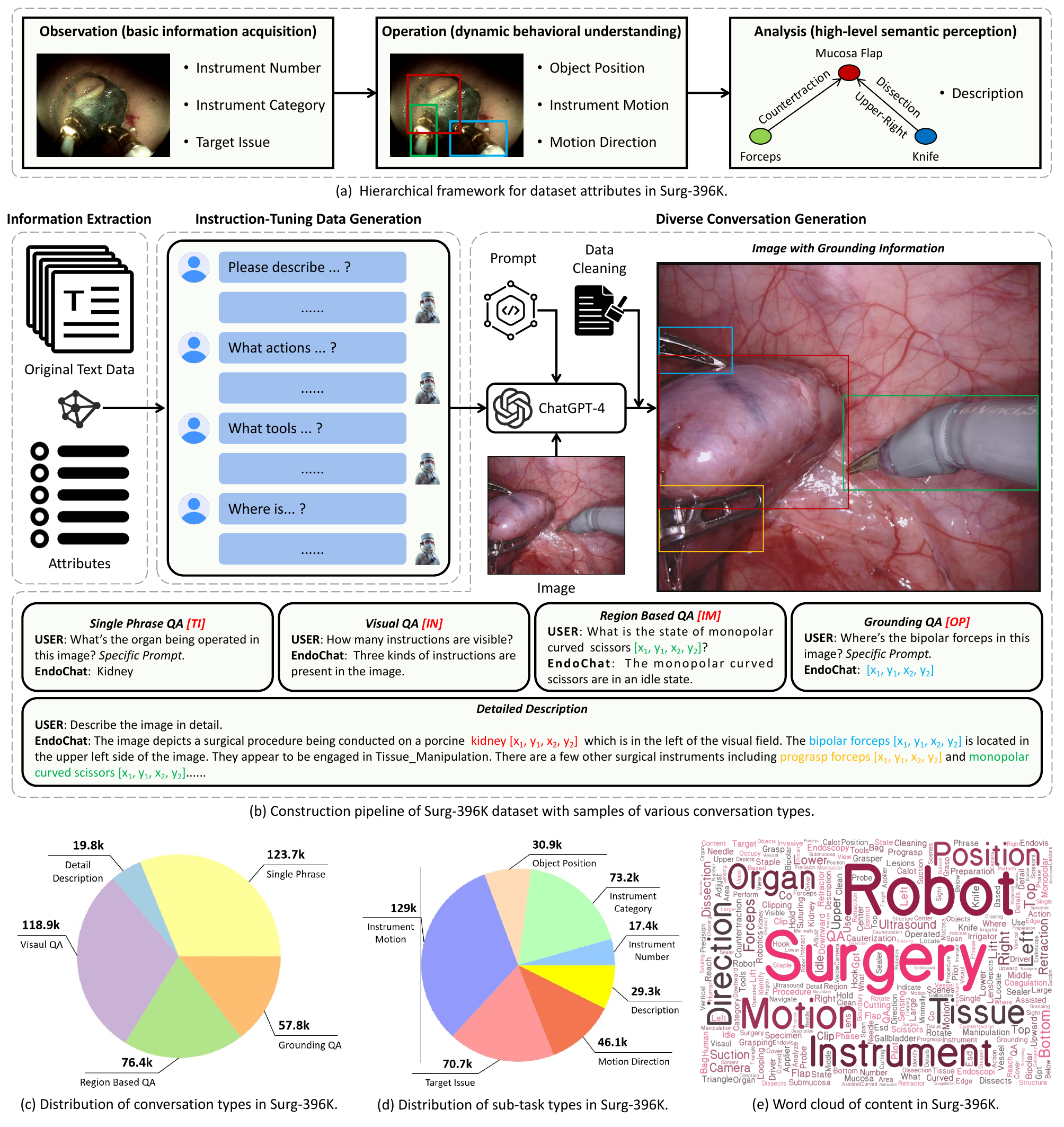}
    \caption{Overview of the construction pipeline and distribution statistics for our Surg-396K dataset. The pipeline involves five key steps: annotation attribute analysis, information extraction, instruction-tuning data generation, diverse conversation generation, and data cleaning.}
    \label{fig:data}
\end{figure*}

The AI-assisted surgery field has experienced a notable expansion in the availability of public multimodal datasets, particularly VQA pairs, as evidenced by works ranging from \cite{seenivasan2022surgical} to \cite{bai2025surgicalvqla}. However, the availability of multimodal instruction data remains limited, primarily due to the time-intensive and less standardized processes involved in human crowd-sourcing. To promote the development of MLLMs tailored for surgical understanding, we propose Surg-396K, a surgical multimodal instruction dataset incorporating 41K images and 396K instruction-following annotations for endoscopic surgery. Following the data generation process shown in Figure~\ref{fig:data}, we define five conversational paradigms and seven attribute-related surgical sub-tasks to capture the majority of natural language dialogue scenarios while ensuring comprehensive coverage of surgical scene understanding. The style of these conversation types and sub-tasks are shown in Table~\ref{tab:data}.

%Constituent Datasets + 一个表
\subsubsection{Preliminary for Constituent Datasets}
In the construction of our Surg-396K dataset, we integrate three distinct datasets. Inspired by the achievements of recent MLLMs in text-annotation tasks~\cite{liu2024visual}, we utilize GPT-4V to expand multimodal instruction-following data, resulting in five conversation types derived from EndoVis-VQLA~\cite{bai2023surgical} and CoPESD~\cite{wang2024copesd} datasets. The third dataset Cholec80-VQA~\cite{seenivasan2022surgical}, which lacks grounding information, is directly employed in two of the conversation types in Surg-396K dataset.

%Surg-396K encompasses two types of annotation, including: GPT-generated and manually labeled. For the first part, inspired from the achievements of recent MLLMs in text-annotation tasks~\cite{liu2024visual}, we utilize GPT-4V to generate multimodal instruction-following data, resulting in five distinct conversation types derived from EndoVis-17/18-VQLA~\cite{bai2023surgical} and CoPESD~\cite{wang2024copesd} datasets. Each conversation type focus on various aspects of image comprehension and question-answering within surgical contexts. Additionally, we introduce Cholec80-VQA~\cite{seenivasan2022surgical}, which is a manually labeled multimodal dataset and directly employed in two of the aforementioned conversation types, further enriching our dataset. 

\noindent \textbf{EndoVis-VQLA}~\cite{bai2023surgical} is a publicly available dataset for endoscopic surgery, derived from the MICCAI Challenges of 2017\cite{allan2019endovis17} and 2018~\cite{allan2020endovis18}. This dataset integrates VQA annotations with bounding box labels to create Visual Question Localized-Answering (VQLA) pairs, which encompass surgical actions, target tissues, instruments, and their respective bounding boxes. The images in EndoVis-VQLA have a resolution of 1280 × 1024 pixels. The dataset is comprised of two parts: EndoVis-18-VQLA, containing 2007 frames, and EndoVis-17-VQLA, including 97 frames.

\noindent \textbf{CoPESD}~\cite{wang2024copesd} is a comprehensive multi-level surgical motion dataset specifically designed for the training of MLLMs in the context of Endoscopic Submucosal Dissection (ESD). It comprises 17,679 images accompanied by detailed motion annotations derived from over 35 hours of ESD videos. The resolution of these images is 1306 × 1009 pixels. The motion annotations include information on target tissues, instruments, surgical motions, motion directions and the corresponding bounding boxes. 

\noindent \textbf{Cholec80-VQA}\cite{seenivasan2022surgical} is an innovative dataset generated from 40 video sequences of the Cholec80 dataset\cite{twinanda2016endonet}, encompassing a total of 21,591 frames. The images in Cholec80-VQA have a resolution of 854 × 480 pixels. Utilizing original tool-operation and phase annotations from the Cholec80 dataset, Cholec80-VQA proposes two types of question-answer pairs for each frame: Classification, which features 14 unique single-word answers; Sentence, which is presented in full sentence form. Due to the absence of grounding information and less content in the annotation of each image, we do not expand it with GPT-4V, but directly use Classification and Sentence as the conversation of the Single Phrase and Visual QA. 

\subsubsection{Attribute Retrieval}
\label{Sec:AE}
%To enhance the accuracy of instruction-following data reference and generation, we have extracted a attribute set that encapsulates critical dimensions of surgical understanding. These attributes, detailed in Table~\ref{tab3}, comprise six distinct categories. \textit{Instrument Category} classifies the types of instruments, facilitating accurate instrument identification. \textit{Instrument Motion} delineates the dynamic functions of each instrument in sequential frames, capturing their operational roles. The attribute \textit{Target Number} quantifies the count of visible instruments per frame.  \textit{Instrument Position} assigns a spatial location to each instrument within a 3×3 grid, providing insights into spatial arrangements. \textit{Target Issue} identifies anatomical targets which are crucial for delineating regions of surgical intervention. Lastly, \textit{Motion Direction} characterizes seven cardinal and diagonal directions (just CoPESD dataset), which is crucial for understanding the instrument motion during surgical procedures. By extracting the attributes of the original datasets, we design specific prompts for each conversation category to fully utilize the content of the original text annotation. 

To ensure that the annotations encompass comprehensive surgical information, we adopt a hierarchical framework for attribute analysis, from basic observation to dynamic operation and high-level perception, as shown in Figure~\ref{fig:data} (a). At the Observation level, foundational attributes are defined, including \textit{Instrument Number} (IN, count of visible instruments), \textit{Instrument Category} (IC, classification of instrument types), and \textit{Target Issue} (TI, anatomical targets of surgical focus). The Operation level focuses on dynamic behaviors and spatial characteristics, such as \textit{Object Position} (OP, spatial mapping within a 3×3 grid), \textit{Instrument Motion} (IM, functional roles inferred from motion), and \textit{Motion Direction} (MD, trajectories across eight cardinal and diagonal directions). Finally, the Analysis level integrates these attributes to comprehensively analyze the surgical scenes, summarized as \textit{Description}. This structured design achieves a seamless integration of the dataset content that maximizes the extraction of surgical information from the original annotations. Furthermore, we develop attribute-specific QA templates tailored for the generation of instruction-tuning data.
%The distribution statistics for these attributes are shown in Figure~\ref{fig:data} (d).   

\begin{table*}[]
\caption{List of attributes (with abbreviations) and conversation types designed for Surg-396K.}
\label{tab:data}
\resizebox{0.96\textwidth}{!}{
\begin{tabular}{c|l|l}
\toprule[1pt]
                                                                             \multicolumn{1}{l|}{}                                                         & Category                     & Response  Style                                                         \\ \hline
\multirow{5}{*}{\begin{tabular}[c]{@{}c@{}}Conversation\\ Type\end{tabular}} & Single Phrase                & Answer the question with a single word or phrase.              \\

                                                                             & Detail Description           & Describe the image in detail with {[}grounding{]}.             \\
                                                                             & Visual QA                    & Question and answer without {[}grounding{]}.                   \\
                                                                             & Region Based QA              & Question with {[}grounding{]}, answer without {[}grounding{]}. \\
                                                                             & Grounding QA                 & Answer the question with a {[}grounding{]}.                    \\ 

                                                                             \midrule

\multicolumn{1}{l|}{}                                                         & Category                     & Sample                                                         \\ \hline
\multirow{8}{*}{\begin{tabular}[c]{@{}c@{}}Sub-task\\ Type\end{tabular}}                                                  & Instrument Number {[}IN{]}   & 2, 3, etc.                                                     \\
                                                                             & Instrument Category {[}IC{]} & Prograsp forceps, ultrasound probe, etc.                       \\
                                                                             & Object Position {[}OP{]} & Right-bottom, left-top, etc.                                   \\
                                                                             & Instrument Motion {[}IM{]}   & Idle, lift, etc.                                               \\
                                                                             & Target Issue {[}TI{]}        & Kidney, the mucosal flap, etc.                                  \\
                                                                             & Motion Direction {[}MD{]}    & Upward, lower left, etc.                                       \\
                                                                             & Description      & Illustrate the image through a descriptive explanation., etc. \\ \bottomrule[1pt]
\end{tabular}}
\end{table*}

\subsubsection{Diverse Conversation Generation}
The expression format of the instruction-tuning data obtained by information extraction through attributes is limited to human-designed templates, resulting in a homogeneous structure. In order to emulate the natural language expression, we diversify the instruction-tuning data by incorporating interaction requirements specific to surgical scenarios, ensuring the coverage of different levels of inquiry and information needs. For example, a trainee might ask a brief question about a specific target, which does not require any redundant information. Therefore, \textit{Single Phrase QA} is designed to provide concise, accurate, and direct responses relevant to surgical contexts. When a trainee requests a detailed explanation of an entire surgical image, \textit{Detailed Description}, on the other hand, offers surgeons comprehensive explanations of all sub-tasks present within the current surgical scene. For routine queries about image content, \textit{Visual QA} delivers context-aware answers by combining user queries with image information. When a trainee needs a focused conversation about a specific region, \textit{Region Based QA} focuses on targeted responses for specific areas of interest, while \textit{Grounding QA} generates bounding box descriptions based on user-provided prompts. Based on this diversification, we interpret the instruction-tuning data through GPT-4V and generate five conversation types:

\noindent \textbf{Single Phrase} is designed to equip EndoChat with the capability to deliver concise, definitive answers to each query, leveraging a rapid analysis of the surgical image. 
%The queries encompass aspects such as instrument types, counts, actions, and their relative positions within the surgical scene, thereby capturing critical elements without extensive explanation. 
This type of conversation can be directly sourced from the instruction-tuning data. Additionally, we have enriched the diversity of the questions utilizing GPT-4V. To guide the model to provide the answer in a single word or phrase, we introduce the task-specific prompt \textit{“Answer the question with a single phrase.”} at the end of the questions, which can be represented by: 
\begin{equation}
\begin{aligned}
\label{eq1}
\mathtt{Human: } \mathbf{T}_{\mathrm{q}}\text{ }[\textit{Prompt}] \backslash\text{n } \mathtt{EndoChat: } \mathbf{T}_{\mathrm{a}}\backslash \text{n}
\end{aligned}
\end{equation}

\noindent \textbf{Detailed Description} provides comprehensive, grounded responses to queries that delve into the intricate details of the visual scene. Answers in this type are entirely generated by GPT-4V with prompts that take full advantage of instruction-tuning data. Therefore, we ensure the response covers all attributes in the surgical images. The list of questions has also been diversified using GPT-4V. This type can augment EndoChat's proficiency in articulating complex visual information as if it is observing the scene in real-time.

\noindent \textbf{Visual QA} emphasizes straightforward question-answer pairs that provide general insights into the surgical scene without specific grounding. This mode differs from Single Phrase by allowing more contextual information in the responses while still maintaining conciseness. Therefore, the generation process for Visual QA entails an additional step compared to Single Phrase. Specifically, GPT-4V is utilized to elaborate the single-word or single-phrase response into a complete sentence while incorporating the content related to description and reasoning attributes from the instruction-tuning data.

\noindent \textbf{Region Based QA} incorporates grounding information within the question to guide the model's attention to a specific region of the image. This conversation type facilitates targeted analysis of visual content, pinpointing the precise location of surgical objects compared to Visual QA. For text expression, we insert the bounding box of the target after its text, e.g., \textit{"kidney [$x_1, y_1, x_2, y_2$]"}. $x_1$ and $y_1$ denote the coordinates of the top-left corner of the bounding box, while $x_2$ and $y_2$ specify the bottom-right corner. Each coordinate value is normalized to the interval $[0, 1]$. 

\noindent \textbf{Grounding QA} delivers responses solely through bounding boxes, training EndoChat to provide accurate spatial answers based on the interplay between the visual content and the posed questions. We introduce the task-specific prompt \textit{“Answer the question with just a bounding box.”} at the end of the questions to guide the model to provide the answer in a bounding box. This conversation type has the same format as Single Phrase QA in Equation~\ref{eq1}. 

More details for these conversation templates are provided in the supplementary material.

\subsubsection{Surgical Sub-task Formulation}
While we have established diverse conversation paradigms to enable MLLMs to handle a wide range of natural language dialogue scenarios, it is critical to ensure a comprehensive understanding of surgical scenes. Given that various elements relevant to surgical scenes have already been extracted through prior attribute retrieval, we formulate seven attribute-related sub-tasks to systematically evaluate MLLMs from different aspects of surgical understanding. The QA pairs for most sub-tasks are derived from Single Phrase QA and Grounding QA since Single Phrase QA can provide concise responses capturing fundamental visual attributes such as instrument types, quantities, and basic spatial relationships, whereas those from Grounding QA offer precise spatial annotations that delineate object positions and regions of interest. Additionally, description-related sub-tasks are directly derived from the Detailed Description paradigm. These attribute-driven sub-tasks encompass a broad spectrum of surgical scene understanding, ranging from basic observation to high-level analytical reasoning, facilitating a comprehensive assessment of MLLMs in understanding complex surgical environments.

\subsubsection{Data Cleaning}
Following the generation of diverse conversations, a data cleaning process is implemented to ensure the integrity and reliability of the training data. Given the large scale of the Surg-396K, we conduct the sampling-based inspection with a ratio of 1/5. Specifically, we manually review the sampled text to evaluate its information completeness, relevance, and clarity. Information completeness refers to whether the text includes all essential content, such as operations and tools. We assign conversation type labels to each QA pair and assess relevance by verifying whether the content aligns with its assigned labels. Since the QA pairs are semantically enriched using GPT-4V, we also inspect whether the enriched text is semantically clear and accurate. During the data cleaning process, we document frequently occurring issues and apply modifications to non-sampled content accordingly. This procedure ensures that retained and revised annotations can provide meaningful information for MLLM training.

\subsubsection{Comparison with Existing Surgical Scene Understanding Datasets}

We present a comprehensive comparison between Surg-396K and existing surgical scene understanding datasets, evaluating key factors such as surgery type, dataset scale, and annotation diversity. Table~\ref{tab:com} provides a detailed summary of Surg-396K in relation to both earlier and more recent benchmarks developed for surgical scene understanding tasks.
Surg-396K exhibits a substantial advantage in both scale and annotation diversity compared to existing datasets. It comprises 41.4K images and 396K annotations spanning multiple surgical procedures, including Laparoscopic Cholecystectomy, Nephrectomy, and Submucosal Dissection, significantly surpassing datasets such as Cholec80-VQA and EndoVis-18-VQA in both volume and variety. Furthermore, in contrast to datasets like CoPESD and PSI-AVA-VQA, Surg-396K's extensive annotation set facilitates more comprehensive training for MLLMs. While SSG-VQA contains a larger number of QA pairs, it contains unimodal annotation format and monotonous QA structure. In contrast, Surg-396K integrates diverse conversational formats alongside grounding annotations, enabling a broader range of multimodal tasks. This combination of scale, diversity, and multimodal richness establishes Surg-396K as a more comprehensive and versatile benchmark, advancing research in fine-grained surgical scene understanding.

\begin{table}[]
\centering
\caption{The comparison of Surg-396K with existing surgical scene understanding benchmarks. In the "Surgery Type" column, "LC" indicates Laparoscopic Cholecystectomy, "Ne" indicates Nephrectomny. "Pr" indicates Prostatectomy. "SD" indicates Submucosal Dissection.}
\label{tab:com}
\resizebox{0.85\textwidth}{!}{
\begin{tabular}{c|ccccc}
\toprule[1pt]
Dataset      & Years & Surgery Type & Image Size          & Annotations & Annotation Size             \\ \hline
% Cholec80~\cite{twinanda2016endonet}     & 2016  & LC           & 86K Images    & Surgical Phases; Bbox & 176110 \\
% CholecT50~\cite{nwoye2023cholectriplet2022}    & 2022  & LC           &  100.9K Images & Triplet  &   161K           \\
% HeiChole~\cite{wagner2023comparative}     & 2023  & LC           & 33 Videos     & Surgical Phases; Mask \\ \hline
VQA-Med-2018~\cite{hasan2018overview} & 2018 & \textbackslash & 2.9K& QA Pairs& 6.4K \\

VQA-Med-2019~\cite{ben2019vqa} & 2019 & \textbackslash & 4.2K& QA Pairs& 15.3K \\

Cholec80-VQA~\cite{seenivasan2022surgical} & 2022  & LC           & 21.6K  & QA Pairs        & 43K          \\

EndoVis-18-VQA~\cite{seenivasan2022surgical} & 2022  & Ne           & 2K  & QA Pairs        & 11.8K          \\

EndoVis-VQLA~\cite{bai2023surgical} & 2023  & Ne           & 2.2K   & QA Pairs; Bbox  &9.5K           \\
PSI-AVA-VQA~\cite{seenivasan2023surgicalgpt}  & 2024  & Pr           & 2.2K   & QA Pairs   &     10.3K           \\
CoPESD~\cite{wang2024copesd}       & 2024  & SD           & 17.7K   & QA Pairs; Bbox   & 17.7K         \\
SSG-VQA~\cite{yuan2024advancing} & 2024  &    LC        & 25K  & QA Pairs        & 960K          \\
Surg-396K (Ours)    & 2025  & LC; Ne; SD   & 41.4K   & QA Pairs; Bbox       & 396K      \\ \bottomrule[1pt]
\end{tabular}}
\end{table}

\subsection{Visual Enhanced MLLM: EndoChat}

Visual grounding conversations in endoscopic surgery involve the intricate interaction between visual and linguistic modalities, requiring a comprehensive understanding of knowledge about distinct objects or regions. Therefore, we propose EndoChat, a novel multimodal large language model designed for visual grounding conversations within the endoscopic surgery scenes, as shown in Figure~\ref{fig:main}. 
Given an input image, the mixed visual encoder extracts source tokens, denoted as $X_d \in \mathbb{R}^{N \times D \times L_1}$ and $X_o \in \mathbb{R}^{N \times D \times L_2}$, where $N$ represents the number of frames, $D$ denotes the hidden dimension, and $L_1$ and $L_2$ correspond to the sequence lengths of the respective token sets. Subsequently, the extracted source tokens are processed by our proposed Mixed Visual Token Engine. The resulting enhanced image tokens are represented as $X^{\prime} \in \mathbb{R}^{N \times (D + m) \times (L_1 + L_2)}$, where $m$ denotes the number of newly generated tokens. These enriched visual tokens are then aligned with language tokens and input to the language model to produce the final response. Moreover, a visual contrast mechanism is introduced to mitigate object hallucinations.

\subsubsection{Preliminary for EndoChat}
EndoChat is built upon the SPHINX architecture~\cite{lin2023sphinx}, a versatile multi-modal large language model designed to support a range of visual instruction-following tasks. The architecture of SPHINX builds upon the large language model LLaMA-2\cite{touvron2023llama}, incorporating multiple vision encoders and employing a joint mixing strategy for weights, tasks, and visual embeddings. To enhance its visual understanding, SPHINX mixes visual embeddings from different vision backbones and processes high-resolution images through a novel strategy of dividing the image into multi-scale sub-images. The integration of multi-task training and the joint mixing strategies collectively empower SPHINX with robust multi-modal capabilities, encompassing tasks such as object detection, diagram interpretation and region-level captioning.

% \subsubsection{Overview of EndoChat}

%, which augments the representation by generating additional visual tokens
%Moreover, we introduce a visual contrast mechanism to mitigate object hallucinations, further enhancing the model's visual understanding and ensuring consistency between visual inputs and language outputs in complex endoscopic surgery scenarios.

%is composed of three primary components: (i) mixed visual encoder, (ii) the multi-scale contextual multi-token engine, and (iii) the large language model (iv) hallucination mitigation module. Unlike the original SPHINX, which cannot interpret endoscopic surgery images, our EndoChat is equipped for this capability through our specialized dataset. The components of this architecture are detailed as follows:
%Additionally, we incorporate task-specific prompts that indicates the task type desired from the model. Moreover, u

\subsubsection{Mixed Visual Token Engine}
In our EndoChat, we mix visual embedding to more scales from high-resolution sub-images, thereby enhancing the encoding of high-resolution images. For input images with high resolution, we implement two parallel pathways to generate five corresponding images at resolutions of 224×224 and 512×512, respectively. Then, these images are fed into a mixed visual encoder, which consists of DINOv2~\cite{oquab2023dinov2} and OpenCLIP~\cite{radford2021learning}, resulting in outputs $X_d$ and $X_o$. For MLLMs, visual encoders typically summarize the visual embeddings after encoding image tokens by extracting an aggregated representation through operations like multi-layer perceptron (MLP). Although this direct representation is computationally efficient, it struggles to capture multi-scale information and often overlooks crucial spatial relationships between different positions or regions. Thus, it may confuse the LLM and underutilize its capabilities. To address these limitations, we introduce the Mixed Visual Token Engine (MVTE). MVTE dynamically generates global visual tokens based on the source token produced by the mixed visual encoder, which seamlessly integrates and maximizes the informational utility of multi-scale visual tokens. 

\begin{figure*}
    \centering
    \includegraphics[width=0.96\textwidth, trim=0 0 0 0]{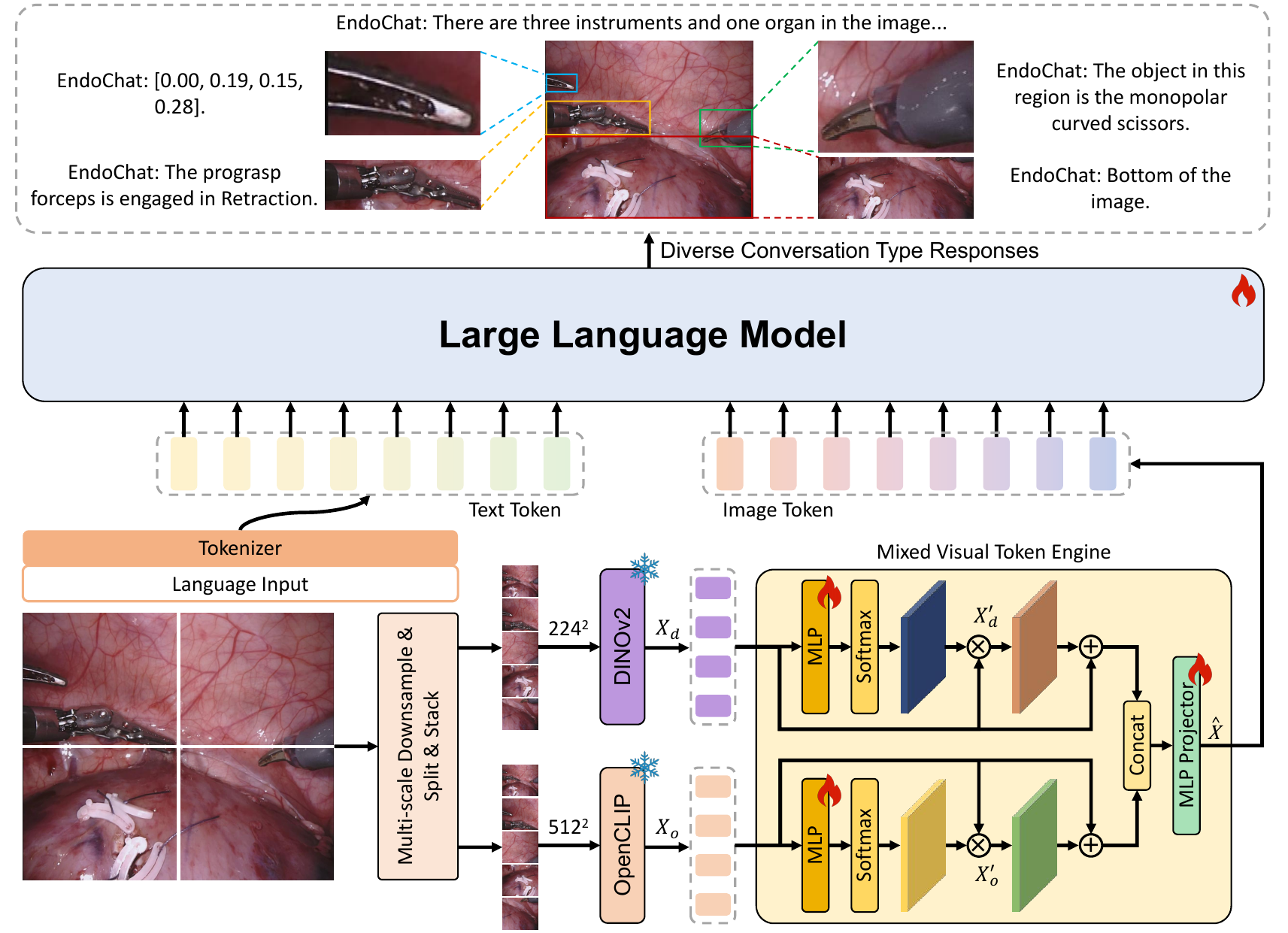}
    \caption{The overview of the proposed EndoChat. For each input image, we use a multi-scale downsampling strategy to generate different scales and sub-images. $224^2$ and $512^2$ indicate concatenated features with the shapes 5×224×224×3 and 5×512×512×3, respectively. These features are subsequently encoded using a mixed visual backbone, followed by the Mixed Visual Token Engine. The resulting vision tokens are then transformed into language space, suitable for input to the Large Language Model. In addition to visual inputs, region coordinates can be auxiliary inputs, along with specific prompts to guide user-defined tasks. This enables the LLM to generate language responses for related object regions.}
    \label{fig:main}
\end{figure*}

Specifically, as shown in the right bottom of Figure~\ref{fig:main}, there are two parallel pathways to process source tokens $X_d$ and $X_o$ from the mixed encoder. In each path, a contextual MLP network (Linear-ReLU-Linear) followed by Softmax normalization is employed to generate the contextual attention map~\cite{song2024pneumollm}. Subsequently, we utilize matrix multiplication to compute the output visual tokens which are spatial-wisely concatenated with their source token to obtain the combined tokens $X^{\prime}$:

\begin{equation}
\begin{aligned}
\label{eq2}
X^{\prime} = softmax(MLP(X)) \cdot X \oplus X
\end{aligned}
\end{equation}
Finally, we channel-wisely concatenate two pathways' combined tokens: $X_o^{\prime}$ and $X_d^{\prime}$, followed by an MLP Projector for dimension alignment to obtain the final image tokens $\hat{X}$. The process can be described in the following equation:

\begin{equation}
\begin{aligned}
\label{eq3}
\hat{X} = Proj(X_o^{\prime} \oplus X_d^{\prime})
\end{aligned}
\end{equation}
The inclusion of MVTE enables the LLM to generate more complementary features, thereby enhancing its comprehension of complex endoscopic surgical scenes and improving effectiveness in complex reasoning tasks.
%thereby enhancing global-local information fusion. The incorporation of MVTE provides more nuanced guidance for the LLM in decoding multi-scale visual features, significantly advancing the model's effectiveness in complex reasoning tasks.

%\subsubsection{Task Prompting}

\subsubsection{Hallucination Mitigation through Visual Contrast}

The MLLMs, parameterized by $\theta$, are adept at capturing intricate visual patterns $x$ and textual query $q$, translating them into coherent linguistic representations $y$. Specifically, MLLMs sample the response $y$ auto-regressively from the probability distribution, predicting the next word step by step based on $x$ and $q$, expressed as: 
\begin{equation}
\begin{aligned}
y_{t} & \sim p_{\theta}\left(y_{t} \mid x, q, y_{<t}\right), \propto \exp \operatorname{logit}_{\theta}\left(y_{t} \mid x, q, y_{<t}\right)
\end{aligned}
\end{equation}
where $y_{t}$ represents the token at time step $t$, and  $y_{<t}$  denotes the sequence of generated tokens up to time step $t-1$. In challenging visual scenarios like endoscopic surgery, MLLMs suffer from Object Hallucination, a phenomenon that arises from their reliance on statistical biases and unimodal priors. This dependency leads to generated text that, while semantically coherent, can be inconsistent with the objects in a given image. Due to the complexity of the endoscopic surgery scenario, ambiguous visual features can lead the MLLMs to ignore critical visual cues, instead favoring linguistic priors in natural pretraining datasets when generating outputs.

%In challenging visual scenarios, such as endoscopic surgery, MLLMs suffers from Object Hallucination. Such phenomenon occurs duo to MLLMs' dependency on statistical biases and unimodal priors, which results in generated text that is semantically coherent but inconsistent with the objects in a given image. Due to the complexity of the endoscopic surgery scenario, ambiguous visual features can lead the MLLM to ignore critical visual cues, instead favoring linguistic priors in natural pretraining datasets when generating outputs, as the LLM is designed to predict the probability of the next word based on a large text corpus.

To address object hallucinations within MLLMs, we introduce the contrast of the model's output generated based on the original and distorted visual input to counteract the statistical biases and language priors~\cite{leng2024mitigating}. Visual contrast is a training-free approach that is grounded in generating two parallel output distributions: one based on the original visual input $x$ and another based on a distorted version $x^{\prime}$. The distorted input $x^{\prime}$ is produced by applying controlled Gaussian noise to $x$, which amplifies language priors and statistical biases that contribute to hallucinations. The contrastive probability distribution $p$ is computed through the logit differences between the original and distorted visual inputs as follows:

\begin{equation}
\begin{aligned}
p(y | x, x', q) = \text{softmax} \Big[ & (1 + \alpha) \cdot \text{logit}_{\theta}(y | x, q) - \alpha \cdot \text{logit}_{\theta}(y | x', q) \Big]
\end{aligned}
\end{equation}
where $\alpha$ is a hyperparameter that adjusts the weighting between the two distributions, with higher $\alpha$ values enhancing the distinction between the two distributions. Such visual contrast serves as a corrective mechanism, reducing hallucinations by contrasting against a distribution predisposed to favoring them. Furthermore, to prevent $p$ from punishing valid outputs and facilitate the generation of a correct token, an adaptive constraint~\cite{li2022contrastive} is introduced:

\begin{equation}
\begin{aligned}
\mathcal{L}\left(y_{<t}\right)=\left\{y_{t} \in \mathcal{L}: p_{\theta}\left(y_{t} \mid x, q, y_{<t}\right) \geq \beta \max _{w} p_{\theta}\left(w \mid x, q, y_{<t}\right)\right\}, \\
p\left(y_{t} \mid x, x^{\prime}, q\right)=0, \text { if } y_{t} \notin \mathcal{L}\left(y_{<t}\right)
\end{aligned}
\end{equation}
where $\beta\in [0,1]$ controls the truncation of the next token distribution. Larger $\beta$ indicates more aggressive truncation, keeping only high-probability tokens. Incorporating adaptive plausibility constraints refines the contrastive distribution, enhancing confidence in decisions. This streamlines the candidate pool, often retaining a single high-probability token, and neutralizes potential adverse effects of visual contrast, preventing the generation of implausible tokens and preserving content integrity.

\subsubsection{Training Strategy}
We adopt the open-source LLaMA2-13B~\cite{touvron2023llama} large language model as our EndoChat's foundational component. LLaMA2-13B serves as a unified interface for diverse vision-language tasks. To ensure the model’s responses are aligned and contextually effective, task-specific prompts are appended to input data, which helps guide the LLM’s responses. For LLM fine-tuning, we employ the Low-Rank Adaptation (LoRA)~\cite{hu2021lora} technique that introduces two smaller matrices as a low-rank approximation of the large, original matrix. We optimize parameters of low-rank matrices instead of all parameters in the LLM. This adaptation method reduces training time and computational overhead. At the same time, it preserves the model's broader knowledge of generic object categories and spatial landmarks, thereby enhancing its vision-language reasoning capabilities in the endoscopic surgery domain.

\section{Results}
\subsection{Implementation Details}
We conduct comparative experiments against multiple MLLMs, including BiomedGPT~\cite{zhang2023biomedgpt}, LLAVA-Med~\cite{li2024llavamed}, Qwen2-VL~\cite{wang2024qwen2}, MiniGPTv2~\cite{chen2023minigpt}, LLAVA-1.5~\cite{liu2024improved}, and SPHINX~\cite{lin2023sphinx}. Additionally, we benchmark our method against a range of specialized models, namely VisualBert~\cite{li2019visualbert}, VisualBert ResMLP~\cite{seenivasan2022surgical}, MCAN~\cite{ben2017mutan}, VQA-DeiT~\cite{touvron2021training}, MUTAN~\cite{ben2017mutan}, MFH~\cite{yu2018beyond}, BlockTucker~\cite{ben2019block}, CAT-ViL DeiT~\cite{bai2023catvil}, GVLE-LViT~\cite{bai2023surgical}, and Surgical-LVLM~\cite{wang2024surgicallvlm}. To train the LLM and the Mixed Visual Token Engine, we utilize an input resolution of 1024×1024. We conduct the training on the Surg-396K dataset for a single epoch, employing four NVIDIA A800 GPUs. We utilize AdamW~\cite{loshchilov2017decoupled} optimizer with an initial learning rate of 2e-5, following a cosine decay schedule and a linear warm-up phase. The training process employs a batch size of 16 and is completed in approximately 20 hours.

%Surg-396K encompasses various conversational paradigms and a wealth of surgical information, and is proposed to augment EndoChat’s ability to deliver comprehensive surgical guidance and interactions.

%enabling users to engage effectively with surgical scene understanding and enhancing surgical education and guidance.

\subsection{EndoChat tailors interactions to varying complexity levels of surgical conversation.}

% \subsection{EndoChat tailors interactions to varying levels of complexity}
% Single Phrase & Detailed Descriptions

EndoChat adapts its interactions to meet the diverse complexities of surgical scenarios through two significant paradigms: Single Phrase QA and Detailed Description. These complementary approaches enable EndoChat to provide precise, actionable insights for both real-time surgical guidance and in-depth educational purposes.

Single Phrase QA focuses on delivering concise and definitive answers, which is ideal for straightforward queries during surgical procedures. By employing the task-specific prompt: ''\textit{Answer the question with a single phrase.}'', EndoChat provides succinct responses, including key aspects such as instrument types, counts, actions, or relative positions within the surgical scene. This capability relies on rapid analysis of visual content, ensuring efficiency without redundant elaboration. For instance, queries like ``How many instruments are visible?'' are answered directly, such as ``three.'' Empirical evaluations, shown in Table~\ref{tab:sin}, demonstrate that EndoChat outperforms the state-of-the-art MLLMs such as BiomedGPT and LLaVA-Med. Specifically, on the EndoVis-17 part, where other models fail to answer questions effectively (0\% accuracy and F-score), EndoChat achieves a remarkable 55.51\% accuracy, along with F-score (29.78), AP@50 (90.25), and mIoU (86.62). It also has superior performance on other parts of Surg-396K, and other public surgery datasets shown in Table~\ref{tab:endovis-vqla} and \ref{tab:phraseqa_vqa}, which demonstrates the robustness and effectiveness of the instruction-tuning process on our Surg-396K dataset.

Detailed Description caters to more complex scenarios where comprehensive understanding is necessary. This interaction type provides in-depth explanations grounded in the visual content of the surgical scene, making it essential for training scenarios and complex procedures. Answers generated by EndoChat offer detailed insights into tissues, instruments, motions, and other surgical elements, aiding in decision-making and contextual understanding, and also providing the reasoning to justify the answers. As shown in Table~\ref{tab:vqa_regionqa_detailed}, EndoChat’s performance in generating detailed descriptions is rigorously assessed using GPT-4 Score, where it significantly outperforms all MLLMs across all parts of Surg-396K. This demonstrates EndoChat’s advantage in generating detailed, context-aware, and high-quality descriptions.

%each capturing a distinct dimension of output quality. BLEU-4 measures n-gram precision, evaluating how well the generated text aligns syntactically with reference descriptions. Specifically, on the EndoVis-18 dataset, EndoChat achieves a BLEU-4 score of 52.20\%, far surpassing models like BiomedGPT (6.59\%) and LLAVA-Med (13.54\%). Regarding CIDEr, EndoChat also leads with a score of 5.9904, a marked improvement over the next best model, LLAVA-Med (1.1236), indicating its proficiency in generating contextually relevant and semantically rich descriptions. Additionally, EndoChat's METEOR score of 40.11\% outshines other models, with LLAVA-Med and BiomedGPT scoring much lower at 20.44\% and 13.07\%, respectively. These results demonstrate EndoChat's comprehensive advantage in generating detailed, context-aware, and high-quality descriptions.

\begin{table*}[h!]
\centering
\caption{Comparison experiments with zero-shot MLLMs in Single Phrase QA and Grounding QA on three parts of Surg-396K dataset.}
\label{tab:sin}
\resizebox{\textwidth}{!}{%
\begin{tabular}{c|cccc|cccc|cccc}
\toprule[1pt]
\multirow{2}{*}{Model} & \multicolumn{4}{c|}{EndoVis-18} & \multicolumn{4}{c|}{EndoVis-17} & \multicolumn{4}{c}{CoPESD} \\ \cline{2-13} 
 & Acc & F-score & AP@50 & mIoU & Acc & F-score & AP@50 & mIoU & Acc & F-score & AP@50 & mIoU \\ \hline
BiomedGPT~\cite{zhang2023biomedgpt} & 5.61 & 3.42 & 39.96 & 32.35 & 0.00 & 0.00 & 38.65 & 36.55 & 5.44 & 1.39 & 36.70 & 28.81 \\
LLAVA-Med~\cite{li2024llavamed} & 3.55 & 1.96 & 32.22 & 28.28 & 0.00 & 0.00 & 28.39 & 17.49 & 8.39 & 3.78 & 54.01 & 51.49 \\
Qwen2-VL~\cite{wang2024qwen2} & 1.99 & 0.22 & 42.13 & 35.35 & 0.00 & 0.00 & 44.49 & 37.82 & 16.72 & 6.29 & 63.59 & 57.59 \\
MiniGPTv2~\cite{chen2023minigpt} & 0.00 & 0.06 & 12.06 & 10.05 & 0.00 & 0.00 & 15.02 & 9.77 & 3.37 & 0.30 & 37.89 & 33.36 \\
LLAVA-1.5~\cite{liu2024improved} & 2.31 & 1.09 & 33.98 & 30.04 & 0.00 & 0.00 & 27.63 & 18.09 & 8.19 & 4.53 & 54.98 & 50.28 \\
SPHINX~\cite{lin2023sphinx} & 1.37 & 0.14 & 32.59 & 27.23 & 0.00 & 0.00 & 25.76 & 15.44 & 7.19 & 4.30 & 59.93 & 55.52 \\
EndoChat & \textbf{71.47} & \textbf{43.74} & \textbf{93.22} & \textbf{86.89} & \textbf{55.51} & \textbf{29.78} & \textbf{90.25} & \textbf{86.62} & \textbf{75.34} & \textbf{31.18} & \textbf{99.43} & \textbf{93.64} \\ \bottomrule[1pt]
\end{tabular}%
}
\end{table*}

\begin{table*}[t]
\centering
\caption{Comparison experiments with zero-shot MLLMs (top) and specialized models (middle) in Single Phrase QA and Grounding QA on EndoVis-VQLA~\cite{bai2023surgical} dataset.}
\label{tab:endovis-vqla}
\resizebox{0.7\textwidth}{!}{%
\begin{tabular}{c|ccc|ccc}
\toprule[1pt]
\multirow{2}{*}{Model}            & \multicolumn{3}{c|}{EndoVis-18-VQLA}                 & \multicolumn{3}{c}{EndoVis-17-VQLA}                 \\ \cline{2-7} 
                          & Acc   & F-score & mIoU & Acc   & F-score & mIoU \\ \hline
BiomedGPT~\cite{zhang2023biomedgpt}                 & 5.61  & 3.42    & 56.93 & 0.00  & 0.00    & 50.59 \\
LLAVA-Med~\cite{li2024llavamed}                 & 3.55  & 1.96    & 53.28 & 0.00  & 0.00    & 49.49 \\
Qwen2-VL~\cite{wang2024qwen2}                  & 1.99  & 0.22    & 53.50 & 0.00  & 0.00    & 47.80 \\
MiniGPTv2~\cite{chen2023minigpt}                 & 0.00  & 0.06    & 26.18 & 0.00  & 0.00    & 22.97 \\
LLAVA-1.5~\cite{liu2024improved}                 & 2.31  & 1.09    & 45.04 & 0.00  & 0.00    & 48.09 \\
SPHINX~\cite{lin2023sphinx}                    & 1.37  & 0.14    & 47.23 & 0.00  & 0.00    & 44.40 \\ \hline
VisualBert~\cite{li2019visualbert}                & 62.68 & 33.29   & 73.91 & 40.05 & 33.81   & 70.73 \\
VisualBert ResMLP~\cite{seenivasan2022surgical}         & 63.01 & 33.90   & 73.52 & 41.90 & 33.70   & 71.37 \\
MCAN~\cite{ben2017mutan}                      & 62.85 & 33.38   & 75.26 & 41.37 & 29.32   & 70.29 \\
VQA-DeiT~\cite{touvron2021training}                  & 61.04 & 31.56   & 73.41 & 37.97 & 28.58   & 69.09 \\
MUTAN~\cite{ben2017mutan}                     & 62.83 & 33.95   & 76.39 & 42.42 & 34.82   & 72.18 \\
MFH~\cite{yu2018beyond}                       & 62.83 & 32.54   & 75.92 & 41.03 & 35.00   & 72.16 \\
BlockTucker~\cite{ben2019block}               & 62.01 & 32.86   & 76.53 & 42.21 & 35.15   & 72.88 \\
CAT-ViL DeiT~\cite{bai2023catvil}              & 64.52 & 33.21   & 77.05 & 44.91 & 36.22   & 73.22 \\
GVLE-LViT~\cite{bai2023surgical}                 & 66.59 & 36.14   & 76.25 & 45.76 & 24.89   & 72.75 \\
Surgical-LVLM~\cite{wang2024surgicallvlm}             & 69.47 & 33.25   & 84.16 & 40.68 & 34.12   & 78.25 \\ \hline
EndoChat          & \textbf{71.47} & \textbf{43.74} & \textbf{86.89} & \textbf{55.51} & \textbf{29.78} & \textbf{86.62} \\ \bottomrule[1pt]
\end{tabular}%
}
\end{table*}

\subsection{EndoChat enhances interactions through multiple modalities}
%EndoChat interacts with humans beyond language

EndoChat also enhances surgical educational interactions with Grounding QA and Region Based QA. Grounding QA outputs bounding boxes to accurately localize surgical elements, providing context-aware guidance for trainees. Region Based QA, on the other hand, uses input bounding boxes to focus on specific areas, such as tools or tissues, aiding in precise localization for hands-on training and surgical navigation.

EndoChat demonstrates superior performance in Grounding QA, a crucial capability for real-time surgical guidance. By delivering responses through bounding boxes, Grounding QA enables EndoChat to ensure accurate spatial localization based on both the visual content and the posed questions. The task-specific prompt, ''\textit{Answer the question with just a bounding box.}'', directs the model to focus on precise spatial information. As shown in Table~\ref{tab:sin} and \ref{tab:endovis-vqla}, EndoChat outperforms other state-of-the-art models by a significant margin. Notably, EndoChat achieves the highest mIoU on both the EndoVis-18-VQLA and EndoVis-17-VQLA datasets, with scores of 86.89 and 86.62, respectively. The consistently superior performance of EndoChat over specialized models highlights its more effective integration of both visual and language-based reasoning, suggesting EndoChat is better suited for surgical navigation tasks.

%For example, LLAVA-Med, an instruction-tuned model specifically designed for medical tasks, achieves an accuracy of 49.6\%, a F-score of 36.2\%, and a mIoU of 75.8\%. While LLAVA-Med benefits from its medical-focused instruction tuning, EndoChat surpasses it with an accuracy of 71.47\%, a F-score of 43.74\%, and a mIoU of 86.89\%. This performance gap highlights EndoChat's more effective integration of both visual and language-based reasoning, suggesting it is better suited for real-time surgical navigation tasks.

Region Based QA, another key interaction, enables EndoChat to provide more targeted analysis by guiding its attention to specific areas within a surgical image. By incorporating the bounding box from the user into the question, Region Based QA helps focus the model on regions of particular interest, such as a surgical tool or tissue. This approach is essential for tasks requiring precise localization or assessment of anatomical structures. As shown in Table~\ref{tab:vqa_regionqa_detailed}, EndoChat outperforms the other models in Region Based QA, achieving the highest on different downstream surgical datasets. This highlights its ability to accurately focus on and analyze localized areas, providing valuable insights for surgical navigation, such as tracking instruments or assessing tissue conditions during procedures. 

\begin{table*}[t]
\centering
\caption{Comparison experiments with zero-shot MLLMs (top) and specialized models (middle) in Single Phrase QA and Visual QA on Cholec80-VQA~\cite{seenivasan2022surgical} dataset.}
\label{tab:phraseqa_vqa}
\resizebox{0.75\textwidth}{!}{%
\begin{tabular}{c|cc|cccc}
\toprule[1pt]
\multirow{2}{*}{Model}         & \multicolumn{2}{c|}{Single Phrase QA} & \multicolumn{4}{c}{Visual QA} \\
\cline{2-7}
                      & Acc   & F-score  & BLEU-3  & BLEU-4  & CIDEr   & METEOR  \\ \hline
BiomedGPT~\cite{zhang2023biomedgpt}             & 8.23  & 3.37    & 5.80    & 2.58    & 0.0159  & 19.62   \\
LLAVA-Med~\cite{li2024llavamed}             & 10.05 & 4.09    & 13.30   & 10.54   & 0.1115  & 30.32   \\
Qwen2-VL~\cite{wang2024qwen2}              & 12.32 & 5.73    & 3.67    & 1.99    & 0.0005  & 18.62   \\
MiniGPTv2~\cite{chen2023minigpt}             & 0.00  & 0.00    & 1.57    & 1.03    & 0.0107  & 7.56    \\
LLAVA-1.5~\cite{liu2024improved}             & 9.99  & 5.52    & 8.44    & 5.90    & 0.0656  & 18.10   \\
SPHINX~\cite{lin2023sphinx}                & 11.67 & 4.12    & 5.18    & 1.13    & 0.0741  & 19.30   \\ \hline
MedFuse~\cite{sharma2021medfusenet}               & 86.10 & 30.90   & 37.80   & 33.30   & 1.2501  & 22.20   \\
VisualBert~\cite{li2019visualbert}             & 89.70 & 63.30   & 96.30   & 95.60   & 8.8020  & 71.90   \\
VisualBert ResMLP~\cite{seenivasan2022surgical}     & 89.80 & \textbf{63.40} & 96.00   & 95.20   & 8.7592  & 71.10   \\
Surgical-LVLM~\cite{wang2024surgicallvlm}         & 87.53 & 60.10   & 96.00   & 95.13   & 8.7755  & 70.88   \\ \hline
EndoChat      & \textbf{92.05} & 61.64   & \textbf{97.28} & \textbf{96.81} & \textbf{9.6702} & \textbf{72.16} \\
\bottomrule[1pt]
\end{tabular}}
\end{table*}

\begin{table*}[t]
\centering
\caption{Comparison experiments with zero-shot MLLMs in Visual QA, Region Based QA, and Detailed Description on Surg-396K dataset.}
\label{tab:vqa_regionqa_detailed}
\resizebox{\textwidth}{!}{%
\begin{tabular}{c|c|ccccc|ccccc|c}
\toprule[1pt]
\multirow{2}{*}{Dataset} & \multirow{2}{*}{Model} & \multicolumn{5}{c|}{Visual QA} & \multicolumn{5}{c|}{Region Based QA} & Detailed Description \\
\cline{3-13}
& & BLEU-4 & CIDEr & METEOR & ROUGE-1 & ROUGE-L & BLEU-4 & CIDEr & METEOR & ROUGE-1 & ROUGE-L & GPT-4 Score \\  \hline
\multirow{7}{*}{EndoVis-18}
& BiomedGPT~\cite{zhang2023biomedgpt}     & 6.59  & 0.7301 & 13.07 & 37.17 & 26.39 & 2.20  & 0.1073 & 13.35 & 28.41 & 27.66 & 38.14 \\
& LLAVA-Med~\cite{li2024llavamed}     & 13.54 & 1.1236 & 20.44 & 54.92 & 36.16 & 4.70  & 0.1595 & 17.35 & 37.23 & 35.84 & 46.40 \\
& Qwen2-VL~\cite{wang2024qwen2}      & 2.39  & 0.5803 & 11.71 & 50.72 & 43.73 & 4.09  & 0.2132 & 17.27 & 43.64 & 39.87 & 46.03 \\
& MiniGPTv2~\cite{chen2023minigpt}     & 1.05  & 0.0235 & 12.89 & 52.33 & 25.25 & 0.88  & 0.0157 & 8.09  & 6.56  & 2.21  & 18.03 \\
& LLAVA-1.5~\cite{liu2024improved}     & 4.91  & 0.3627 & 15.93 & 41.21 & 32.94 & 3.19  & 0.2008 & 16.88 & 42.09 & 39.18 & 25.49 \\
& SPHINX~\cite{lin2023sphinx}        & 15.11 & 0.7862 & 15.53 & 32.18 & 30.14 & 2.57  & 0.1024 & 5.38  & 6.41  & 5.83  & 43.99 \\
& EndoChat & \textbf{52.20} & \textbf{5.9904} & \textbf{40.11} & \textbf{81.20} & \textbf{79.62} & \textbf{59.65} & \textbf{5.5735} & \textbf{41.05} & \textbf{82.01} & \textbf{81.21} & \textbf{79.35} \\
\hline
\multirow{7}{*}{EndoVis-17}
& BiomedGPT~\cite{zhang2023biomedgpt}     & 8.81  & 0.6362 & 15.16 & 40.61 & 32.34 & 3.57  & 0.1874 & 5.99  & 25.74 & 24.02 & 56.06 \\
& LLAVA-Med~\cite{li2024llavamed}     & 12.92 & 0.8814 & 17.13 & 43.71 & 36.36 & 9.03  & 0.3583 & 17.27 & 42.07 & 37.18 & 64.71 \\
& Qwen2-VL~\cite{wang2024qwen2}      & 10.97 & 1.1381 & 20.26 & 43.69 & 42.71 & 7.26  & 0.4505 & 19.42 & 45.22 & 37.93 & 67.57 \\
& MiniGPTv2~\cite{chen2023minigpt}     & 1.80  & 0.0114 & 12.89 & 39.08 & 18.90 & 1.08  & 0.0256 & 9.80  & 10.16 & 9.36  & 29.35 \\
& LLAVA-1.5~\cite{liu2024improved}     & 13.73 & 0.7942 & 17.78 & 47.23 & 37.38 & 8.72  & 0.5145 & 18.09 & 46.36 & 41.06 & 41.75 \\
& SPHINX~\cite{lin2023sphinx}        & 14.12 & 1.2109 & 16.74 & 42.15 & 37.32 & 3.34  & 0.0938 & 6.96  & 16.12 & 13.31 & 54.58 \\
& EndoChat & \textbf{21.75} & \textbf{1.5083} & \textbf{23.41} & \textbf{52.05} & \textbf{46.65} & \textbf{18.12} & \textbf{1.4149} & \textbf{21.25} & \textbf{48.34} & \textbf{43.91} & \textbf{68.67} \\
\hline
\multirow{7}{*}{CoPESD}
& BiomedGPT~\cite{zhang2023biomedgpt}     & 1.62  & 0.0064 & 5.88  & 19.23 & 16.25 & 1.69  & 0.0173 & 7.17  & 25.38 & 22.46 & 34.31 \\
& LLAVA-Med~\cite{li2024llavamed}     & 4.56  & 0.2133 & 14.08 & 42.78 & 35.37 & 6.68  & 0.1489 & 17.42 & 50.70 & 44.04 & 71.10 \\
& Qwen2-VL~\cite{wang2024qwen2}      & 3.03  & 0.3411 & 14.51 & 48.81 & 38.20 & 4.69  & 0.0545 & 16.88 & 54.14 & 45.13 & 51.54 \\
& MiniGPTv2~\cite{chen2023minigpt}     & 2.45  & 0.0936 & 15.03 & 37.92 & 35.45 & 3.75  & 0.0055 & 8.16  & 29.36 & 31.25 & 27.16 \\
& LLAVA-1.5~\cite{liu2024improved}     & 4.51  & 0.1774 & 14.99 & 45.98 & 35.74 & 6.14  & 0.1594 & 18.01 & 52.23 & 45.31 & 49.00 \\
& SPHINX~\cite{lin2023sphinx}        & 7.03  & 0.2601 & 14.98 & 42.30 & 34.75 & 6.19  & 0.0194 & 2.53  & 5.58  & 5.01  & 38.85 \\
& EndoChat & \textbf{46.94} & \textbf{3.2134} & \textbf{39.61} & \textbf{73.56} & \textbf{66.79} & \textbf{49.79} & \textbf{3.4410} & \textbf{38.04} & \textbf{71.98} & \textbf{65.44} & \textbf{82.48} \\
\bottomrule[1pt]
\end{tabular}%
}
\end{table*}

\subsection{EndoChat is a surgeon-like interaction tool}

% \subsection{EndoChat is a human-like interactive tool}
EndoChat leverages the multi-modal conversational instruction-tuning dataset to become a surgeon-like interaction model through Visual Question Answering, where the system answers general questions about the surgical scene while maintaining a balance between conciseness and contextual clarity. Unlike Single Phrase QA, which provides brief responses, Visual QA allows for more detailed insights, elaborating on key aspects of the image. This approach mimics a more natural conversational flow, enabling EndoChat to deliver informative yet succinct answers, resembling how a human expert would provide both simple and insightful feedback during surgery. 

Table~\ref{tab:phraseqa_vqa} demonstrates that EndoChat significantly outperforms both zero-shot MLLMs and specialized models in Visual QA on the Cholec80-VQA dataset. EndoChat achieves the highest score in all evaluation metrics, surpassing SOTA specialized models VisualBert and VisualBert ResMLP. This highlights EndoChat’s effectiveness in tasks that require both fine-grained understanding and detailed responses. Other than that, EndoChat also exhibits a clear advantage in other parts of Surg-396K, which is presented in Table~\ref{tab:vqa_regionqa_detailed}. Its results highlight a significant enhancement in metrics such as BLEU-4 and METEOR.
%, reflecting its superior ability to integrate fine-grained visual features with contextual language understanding.
Compared to SPHINX and other models, which achieve relatively lower scores, EndoChat demonstrates a more comprehensive capability in extracting and reasoning over multimodal information. EndoChat sets a new benchmark for surgical assistance and educational applications. Its strong performance across these key evaluation metrics solidifies its position as a top choice for clinical and academic use in surgical domains.

%Compared to other models like Surgical-LVLM, which achieves lower scores in these metrics (BLEU-4: 95.13\%, CIDEr: 8.7755, METEOR: 70.88\%).

\subsection{EndoChat advances comprehensive understanding of surgical scenerios}
In order to comprehensively evaluate the capacity of EndoChat in addressing the challenges inherent in surgical environments, we conduct an in-context learning comparison between EndoChat and medical-specialized MLLMs in seven surgical scene understanding tasks on the CoPESD part of the Surg-396K dataset. The CoPESD part is chosen since it encompasses the full range of surgical understanding challenges. The tasks are formulated based on the dataset attributes illustrated in Figure~\ref{fig:data} a, with each attribute defining one or two corresponding tasks~\cite{li2024llava}. These tasks reflect the essential components of surgical scenarios, such as instrument recognition, motion understanding, and issue detection. In-context learning is utilized for its advantage to dynamically adapt to new tasks and queries by leveraging task-specific prompts like "\textit{The answer must be one of the following words or phrases: 'Reach', 'Rotate', 'Grasp', 'Lift', 'Hold', 'Stay idle', 'Dissect'.}". As summarized in Table~\ref{tab:comparison}, EndoChat consistently outperformed other medical-specialized MLLMs, showcasing its adaptability and precision in handling diverse understanding challenges.

Firstly, tasks such as instrument counting and object localization highlight fundamental scene comprehension abilities. While all models achieved reasonable performance in instrument counting, EndoChat leads with an accuracy of 85.14\%, surpassing BiomedGPT (78.84\%) and LLaVA-Med (49.88\%). For object localization, EndoChat’s accuracy (39.88\%) exceeded BiomedGPT by over 26\%, demonstrating its superior integration of spatial and semantic cues. Secondly, in more complex tasks such as motion recognition and direction prediction, EndoChat achieved the highest accuracy and F-scores, indicating its robustness in dynamic and context-sensitive scenarios. These results suggest that EndoChat effectively deciphers intricate spatial relationships within surgical scenes. Furthermore, in instrument category identification and target tissue recognition, EndoChat also achieves excellent performance, surpassing all other models. These findings demonstrate its proficiency in distinguishing both instruments and their associated anatomical targets in complex surgical environments. Overall, these findings highlight EndoChat’s capacity to generalize across diverse task types, making it a reliable tool for real-world surgical training and guidance.

\begin{table*}[ht]
\centering
\caption{Comparison experiments with zero-shot medical-specialized MLLMs in various surgical scene understanding tasks (excluding the Description task which has been shown in Table~\ref{tab:vqa_regionqa_detailed}) on the CoPESD part of Surg-396K dataset. Six types of tasks include: the number of instruments, the object location in the surgical scene (textual form), the current motion of the instrument, the direction of instruction motion, the identification of instruments, instrument detection, the recognition of issues and issue detection.}
\label{tab:comparison}
\resizebox{0.8\textwidth}{!}{
\begin{tabular}{c|cc|cc|cc|cc}
\toprule
\multirow{3}{*}{Model} & \multicolumn{8}{c}{CoPESD} \\
\cline{2-9}
& \multicolumn{2}{c|}{Instrument Number} & \multicolumn{2}{c|}{Object Position} & \multicolumn{2}{c|}{Instrument Motion} & \multicolumn{2}{c|}{Motion Direction} \\
\cline{2-9}
& Acc & F-score & Acc & F-score & Acc & F-score & Acc & F-score \\
\hline
BiomedGPT~\cite{zhang2023biomedgpt} & 78.84 & 24.03 & 12.92 & 9.86 & 13.41 & 5.61 & 7.68 & 3.06 \\
LLAVA-Med~\cite{li2024llavamed} & 49.88 & 14.05 & 26.66 & 17.62 & 17.13 & 7.78 & 14.52 & 4.90 \\
EndoChat & \textbf{85.14} & \textbf{32.57} & \textbf{39.88} & \textbf{17.80} & \textbf{68.53} & \textbf{32.47} & \textbf{43.21} & \textbf{22.64} \\
\hline
& \multicolumn{4}{c|}{Instrument Category} & \multicolumn{4}{c}{Target Tissue} \\
\cline{2-9}
& Acc & F-score & AP@50 & mIoU & Acc & F-score & AP@50 & mIoU \\
\hline
BiomedGPT~\cite{zhang2023biomedgpt} & 20.86 & 15.84 & 27.60 & 22.63 & 56.01 & 25.16 & 25.54 & 20.41 \\
LLAVA-Med~\cite{li2024llavamed} & 64.12 & 42.47 & 62.46 & 58.21 & 86.95 & 46.51 & 38.63 & 36.71 \\
Yolov11~\cite{khanam2024yolov11} & \textbackslash &\textbackslash & 90.29 & 82.63  & \textbackslash &\textbackslash & 96.52 & 85.28 \\
EndoChat & \textbf{91.78} & \textbf{91.77} & \textbf{99.78} & \textbf{93.70} & \textbf{97.54} & \textbf{94.09} & \textbf{98.79} & \textbf{93.52} \\
\bottomrule
\end{tabular}}
\end{table*}

\begin{table*}[ht]
\caption{Ablation study with zero-shot medical-specialized MLLMs on EndoVis-18 part of Surg-396K dataset.}
\label{tab:abl}
\centering
\begin{tabular}{c|cc|cc|c}
\toprule
                                                           & \multicolumn{2}{c|}{Single Phrase} & \multicolumn{2}{c|}{Grounding QA}    & Detail Description                  \\ \cline{2-6}
\multirow{-2}{*}{\begin{tabular}[c]{@{}c@{}}Mixed Visual\\ Token Engine\end{tabular}} & Acc             & F-score         & acc@0.5           & mIoU            & GPT-4 Score \\ \hline
$\times$                                                                              & 66.74\%         & 33.75\%         & 93.03\%           & 86.86\%         & 78.26                             \\
\checkmark                                                             & 71.47\%         & 43.74\%         & 93.22\%           & 86.89\%         & 79.35                             \\ \hline
                                                                                      & \multicolumn{2}{c|}{Visual QA}     & \multicolumn{2}{c|}{Region Based QA} & Detail Description                  \\ \cline{2-6}
\multirow{-2}{*}{\begin{tabular}[c]{@{}c@{}}Hallucination \\ Mitigation\end{tabular}} & CIDEr           & ROUGE-L         & CIDEr             & ROUGE-L         & GPT-4 Score                         \\ \hline
$\times$                                                                              & 6.0068          & 79.47\%         & 5.4069            & 80.83\%         & 77.3853                             \\
\checkmark                                                             & 5.9904          & 79.62\%         & 5.5735            & 81.21\%         & 79.3463  \\
\bottomrule
\end{tabular}
\end{table*}

\subsection{Ablation study for the effectiveness of core modules in EndoChat}
In the ablation study, we evaluate the effectiveness of our proposed modules in EndoChat: Mixed Visual Token Engine and Visual Contrast Hallucination Mitigation, utilizing the EndoVis-18 subset of the Surg-396K dataset. The results are shown in Table~\ref{tab:abl}. For the evaluation of MVTE, we focus on three conversation types: Single Phrase, Grounding QA, and Detail Description, since these are particularly sensitive to the quality of image feature extraction and perception. Specifically, for the Single Phrase, accuracy increases from 66.74\% to 71.47\% and F-score from 33.75\% to 43.74\%, demonstrating that MVTE significantly enhances the model’s ability to generate more accurate phrase-level descriptions. While the improvement in Grounding QA is modest, the increase in
GPT-4 score (from 78.26 to 79.35) suggests that MVTE contributes to refining the model’s reasoning capabilities, particularly for complex surgical scenarios. These results indicate that MVTE strengthens the model’s capacity to capture high-quality image features, thus improving performance in tasks requiring fine-grained image perception.

The Hallucination Mitigation module is evaluated on conversation types that are particularly prone to hallucinations, such as Visual QA, Region Based QA, and Detail Description. These conversation types are susceptible to the model generating irrelevant or inaccurate responses due to the challenge of aligning visual content with textual queries. In Visual QA, this module leads to an increase in CIDEr (from 5.9068 to 5.9904) and ROUGE-L (from 79.47\% to 79.62\%), indicating improvements in the semantic accuracy and contextual relevance of the generated responses. The effect is more pronounced in Region Based QA, where CIDEr increases from 5.4069 to 5.5735 and ROUGE-L rises from 80.83 to 81.21, demonstrating that the hallucination mitigation module significantly improves the model’s ability to produce reliable, region-specific answers. Additionally, the increase in the GPT-4 score (from 77.39 to 79.35) further underscores this module’s contribution to enhancing the model’s overall reasoning capabilities. These findings highlight the critical role of hallucination mitigation in improving performance across visual grounding tasks.

\subsection{Expert evaluation of EndoChat by endoscopists}
To validate EndoChat’s potential in advancing surgical training and education, we conduct an expert evaluation involving 150 endoscopic surgery cases, evaluated by experienced endoscopists from Qilu Hospital. Each surgery case comprised a surgical image along with the corresponding five rounds of conversation. To ensure the evaluation’s comprehensiveness, the five rounds of conversation included a detailed description of the scenario, supplemented by four rounds of randomly selected Visual QA and Region Based QA, which encompass all attributes of the surgical data in Surg-396K. Additionally, the ground-truth to these conversations are provided to assist the endoscopist in assessing EndoChat’s descriptive accuracy, analytical depth, and applicability in training scenarios. 

During the endoscopist evaluation, the conversations generated by EndoChat are presented with the indication that the results are produced by MLLMs. The subsequent process is to assess the usability of EndoChat by comparing its generated outputs with the correct answers. Endoscopists then evaluated the results and assigned scores to each case for the following standards:
\begin{itemize}
    \item EndoChat's description is correct.
    \item EndoChat's analysis is useful.
    \item EndoChat can help surgeons during training.
    \item Willingness to use EndoChat.
\end{itemize}

\begin{figure*}
    \centering
    \includegraphics[width=0.98\textwidth, trim=0 0 0 0]{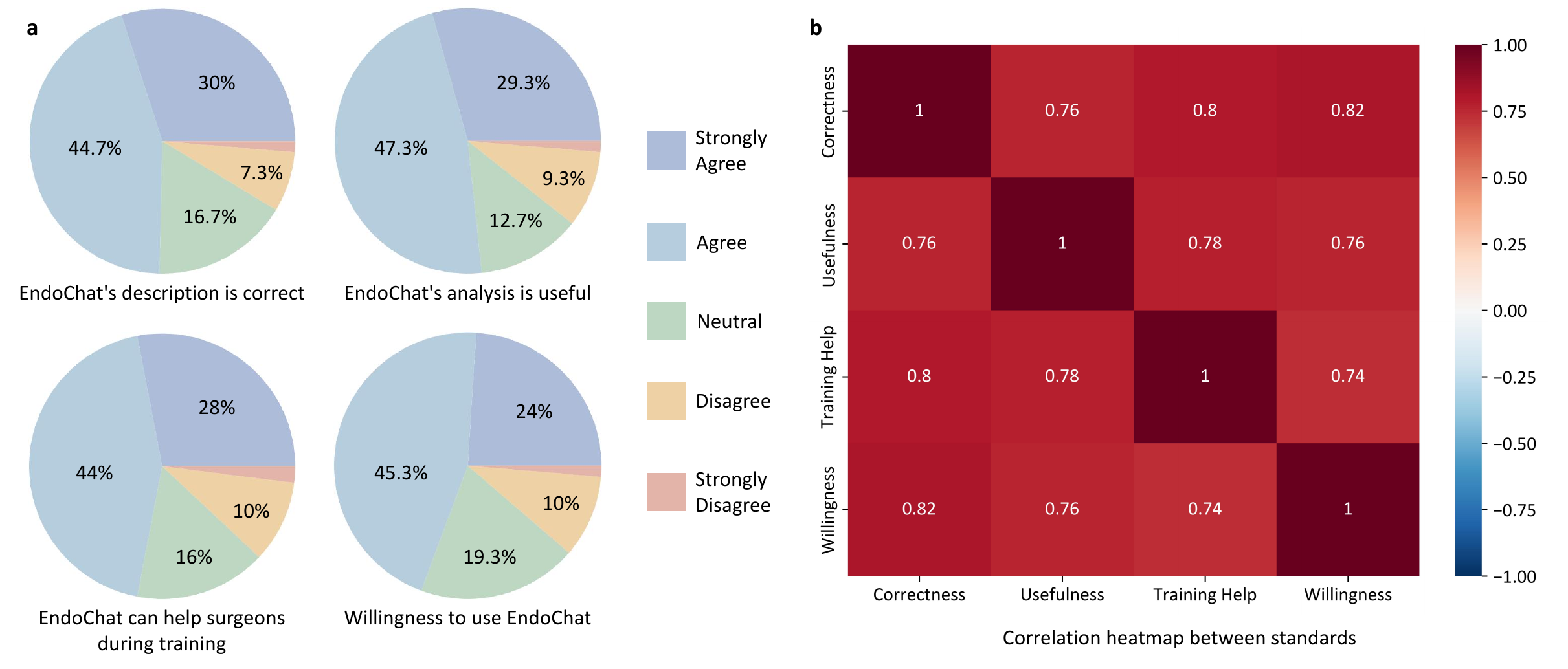}
    \caption{\textbf{Endoscopist evaluation of EndoChat in 150 cases. a} Questionnaire-based evaluation of EndoChat conducted by endoscopists. The pie charts illustrate the distribution of cases in which endoscopists express varying levels of agreement. \textbf{b} Correlation analysis of four evaluation standards.}
    \label{fig:off}
\end{figure*}

As shown in Figure~\ref{fig:off} a, the scores ranged from strongly agree to strongly disagree, and 74.7\% cases are evaluated as having correct descriptions provided by EndoChat, while 76.6\% cases feature useful analysis that enhances the understanding of surgical scenes. Additionally, for 72\% of the cases, endoscopists agreed that EndoChat could effectively assist trainees in surgical training, helping to refine procedural skills and improve educational outcomes. Finally, 69.3\% of the cases reflected a willingness to incorporate EndoChat into surgical training, indicating its potential for real-world adoption. These findings highlight EndoChat’s role as a reliable tool and its effectiveness in advancing surgical training procedures and education in endoscopic surgery. Other than that, Figure~\ref{fig:off}~(b) illustrates the pairwise correlations among the evaluation standards. A strong positive correlation is observed between the correctness of answers and the willingness to use EndoChat, highlighting that its capacity to provide accurate and reliable information directly drives its potential for real-world implementation. Additionally, the training helps exhibit a robust correlation with both the usefulness of analysis and correctness, indicating that practical and insightful analysis is pivotal for acceptance and education. These relationships emphasize that EndoChat's ability to generate precise, contextually relevant outputs aligns with the endoscopists’ need for training support. Given the excellent performance demonstrated in previous experiments, EndoChat shows clear advantages in bridging the gap between AI innovation and endoscopic surgical practice through such correlations, further solidifying its value as a versatile and impactful tool.

\section{Discussion}
This study aims to develop an intelligent surgical chatbot and copilot for surgical education and training. We first construct a high-quality, multi-paradigm dataset, Surg-396K, for surgical scene understanding and dialogue, along with a comprehensive framework for vision-language data collection and annotation in surgical scenarios. Furthermore, we develop the enhanced visual representation extraction and inference strategy, which serves as the foundation for EndoChat, our chatbot system designed to perform multimodal understanding and dialogue in surgical contexts.

Our analysis and comparisons include six MLLMs and over ten specialized models, demonstrating that our model achieved outstanding performance across various dialogue paradigms and surgical-specific scene understanding tasks. We further validate the effectiveness of our proposed visual feature learning approach and the visual contrast-based MLLM reasoning method through ablation experiments. Additionally, we invite experienced endoscopists to evaluate their willingness to use EndoChat as an assistant during training or surgeries. In most evaluation cases, the surgeons provide positive feedback, further showcasing the clinical reliability, usability, and acceptability of our proposed EndoChat. There are demonstrations of qualitative comparison in supplementary material.

Generally, there are two key factors in designing a precise and surgeon-friendly chatbot for dialogue and usage. The first is to ensure accuracy in downstream tasks for surgical scene understanding, such as instrument recognition and action identification~\cite{saab2024capabilities}. To enhance EndoChat's performance in these downstream tasks, we consider single-phrase QA to be the most critical. This is primarily due to the answers in such dialogue data are often simple words or short phrases, making it easier for the model to link visual information with text annotations, thereby achieving higher accuracy in sub-tasks.
The second is making the chatbot better suited to how surgeons use it, which is a main focus of this study. Establishing different dialogue paradigms helps respond to questions from surgeons and trainees in various contexts, and also helps constrain the divergent dialogue tendencies typical of MLLMs. This allows the model to focus more on the questions posed by surgeons and provide relevant answers.

Despite the impressive performance of our EndoChat on various surgical dialogue tasks, it still faces several limitations. First, although we possess a large surgical image database, the number of unique surgical cases included is relatively small. Such a large database facilitates tasks like action and instrument recognition; however, the limited number of cases may hinder the generalization ability of our model when applied to different surgical techniques~\cite{wang2023huatuo, goetz2024generalization}. Expanding the database to include more surgical procedures and cases in the future could significantly enhance the generalizability and applicability of EndoChat across diverse surgical scenarios. 
In addition, MLLMs often rely on substantial computational power, which poses challenges for deployment in resource-constrained edge environments. Existing deployment approaches include developing lightweight MLLMs for deployment on mobile devices or personal computers, or hosting the model in the cloud and enabling communication with mobile/computer terminals~\cite{wang2024cloud,yao2024minicpm}. Finding ways to deploy MLLMs in clinical settings with limited computational resources will remain a significant challenge. 
Lastly, as more and more diverse data are introduced, issues concerning the privacy and ethical use of clinical data need to be carefully studied and reviewed to ensure compliance during their application.

\section{Conclusion}
In this paper, we introduce Surg-396K, a comprehensive surgical multimodal dataset that includes 396K image-instruction pairs across multiple conversation paradigms. Based on Surg-396K, we present a flexible surgical understanding MLLM, EndoChat, designed to integrate various downstream tasks in surgical scene understanding and support different dialogue paradigms that may occur between surgeons and chatbots. Endochat integrates the Mixed Visual Token Engine (MVTE) and the visual contrast-based hallucination mitigation strategy, allowing it to effectively capture high-quality visual features and reduce inaccuracies in generated responses. Extensive experiments demonstrated the effectiveness of our approach, providing a more generalizable solution for understanding the surgical scene. Furthermore, the positive feedback from expert evaluations underscores the model's practical applicability in real-world surgical environments.

Moving forward, we will open-source our model weights, training code, and data to promote the development of multimodal AI systems in the surgical domain. In the future, we will collaborate with surgeons and clinical systems to conduct more rigorous and extensive validations to ensure the safety, reliability, and usability of the dialogue model. We target to integrate EndoChat directly into surgical training or endoscopic surgery systems. By using monitors and voice-based dialogue systems, EndoChat could provide direct assistance to surgeons or trainees.

\section{Acknowledgements}
This work was supported by HK RGC, Collaborative Research Fund (CRF C4026-21GF), General Research Fund (GRF 14203323, GRF 14216022, and GRF 14211420), NSFC/RGC Joint Research Scheme N\_CUHK420/22, Shenzhen-Hong Kong-Macau Technology Research Programme (Type C) STIC Grant 202108233000303, InnoHK program and National Natural Science Foundation of China (No. 82261160396).

\section{Competing interests}
The authors declare no competing interests.

\section{Author contributions statement}
Guankun Wang: Conceptualization, Methodology, Validation, Investigation, Writing – original draft. Long Bai: Conceptualization, Methodology, Investigation, Writing – original draft. Junyi Wang: Validation, Investigation, Dataset Generation, Writing – review and editing. Kun Yuan: Methodology, Writing – original draft. Zhen Li: Endoscopist Evaluation, Writing – review and editing. Tianxu Jiang: Dataset Generation, Writing – review and editing. Xiting He: Software, Writing – review and editing. Jinlin Wu: Computing Resource Support, Writing – review and editing. Zhen Chen: Conceptualization, Writing – review and editing. Zhen Lei: Formal analysis, Writing – review and editing. Hongbin Liu: Formal analysis, Writing – review and editing. Jiazheng Wang: Conceptualization, Writing – review and editing. Fan Zhang: Conceptualization, Writing – review and editing. Nicolas Padoy: Data curation, Writing – review and editing. Nassir Navab: Conceptualization, Data curation, Writing – review and editing. Hongliang Ren: Conceptualization, Supervision, Writing – original draft, Resources.

\bibliographystyle{cas-model2-names}
\bibliography{main}

\end{document}